\documentclass[10pt,a4paper]{article}

\usepackage[T1]{fontenc}
\usepackage[utf8]{inputenc}
\usepackage[a4paper,top=2.3cm,bottom=2.2cm,left=1.9cm,right=1.9cm,headheight=22pt]{geometry}
\usepackage{times}
\usepackage{microtype}
\usepackage{amsmath,amssymb,amsthm}
\usepackage{graphicx}
\usepackage{booktabs,tabularx,multirow,array}
\usepackage{xcolor,colortbl}
\usepackage{tikz,pgfplots}
\usetikzlibrary{arrows.meta,positioning,calc,backgrounds,fit}
\usepgfplotslibrary{groupplots}
\pgfplotsset{compat=1.18}
\usepackage{fancyhdr}
\usepackage{enumitem}
\usepackage{titlesec}
\usepackage{caption}
\usepackage{natbib}
\usepackage{hyperref}
\usepackage{url}

\definecolor{freedoblue}{HTML}{0E4D92}
\definecolor{freedodark}{HTML}{082B54}
\definecolor{freedolight}{HTML}{F4F8FD}
\definecolor{freedoborder}{HTML}{C6D9F1}
\definecolor{poleblue}{HTML}{1676A3}
\definecolor{eventorange}{HTML}{CB6E22}
\definecolor{eventgreen}{HTML}{30836B}
\definecolor{darkgray}{HTML}{222222}
\definecolor{mutedgray}{HTML}{555555}
\definecolor{tabhighlight}{rgb}{0.925,1,1}

\hypersetup{
  colorlinks=true,
  linkcolor=freedoblue,
  citecolor=freedoblue,
  urlcolor=freedoblue,
  pdftitle={Zhengrong-PoleWM: Visual World Models with Pole-Based Action-Response Factorization},
  pdfauthor={Yu Zhang, Jingrui Li, Congyu Wang, Yuhang Cheng, Wenbo Chen, Xiaobo Zhang, Jing Zou, Yuxi Dai}
}

\newcommand{\ours}{\texttt{Zhengrong-PoleWM}}
\newcommand{\R}{\mathbb{R}}
\newcommand{\C}{\mathbb{C}}
\newcommand{\E}{\mathbb{E}}
\newcommand{\MSE}{\operatorname{MSE}}
\newcommand{\sg}{\operatorname{sg}}
\newcommand{\spplus}{\operatorname{softplus}}
\newcommand{\ie}{\emph{i.e., }}

\makeatletter\let\rowinput\@@input\makeatother

\newtheorem{proposition}{Proposition}

\titleformat{\section}
  {\Large\bfseries\color{freedodark}}
  {\thesection}
  {0.6em}
  {}
  [\vspace{2pt}{\color{freedoborder}\titlerule[0.8pt]}]

\titleformat{\subsection}
  {\large\bfseries\color{freedoblue}}
  {\thesubsection}
  {0.5em}
  {}

\setlist[itemize]{leftmargin=*,itemsep=2pt,topsep=3pt}
\setlist[enumerate]{leftmargin=1.6em,itemsep=2pt,topsep=3pt}

\fancypagestyle{firstpage}{
  \fancyhf{}
  \fancyhead[L]{%
    \raisebox{-0.2\height}{\includegraphics[height=15pt]{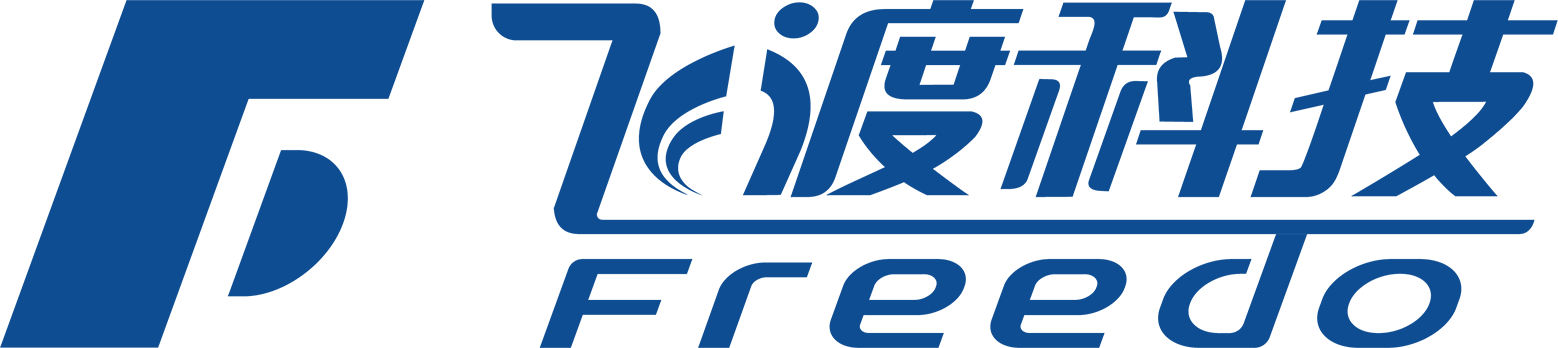}}%
    \hspace{6pt}{\color{freedoborder}\vrule width 0.7pt height 11pt depth 2pt}\hspace{6pt}%
    {\footnotesize\color{mutedgray}\textbf{Beijing Freedo Technology Co., Ltd.}}%
  }
  \fancyhead[R]{{\footnotesize\color{freedoblue}\textbf{FREEDO AI RESEARCH $\cdot$ TECHNICAL REPORT}}}
  \fancyfoot[L]{{\footnotesize\color{mutedgray}Beijing Freedo Technology Co., Ltd.}}
  \fancyfoot[C]{{\footnotesize\color{mutedgray}\ours{} Technical Report}}
  \fancyfoot[R]{{\footnotesize\color{freedodark}\textbf{\thepage}}}
  \renewcommand{\headrulewidth}{0.6pt}
  \renewcommand{\headrule}{\hbox to\headwidth{\color{freedoborder}\leaders\hrule height \headrulewidth\hfill}}
  \renewcommand{\footrulewidth}{0.4pt}
  \renewcommand{\footrule}{\hbox to\headwidth{\color{freedoborder}\leaders\hrule height \footrulewidth\hfill}}
}

\renewcommand{\headrulewidth}{0.5pt}
\renewcommand{\headrule}{\hbox to\headwidth{\color{freedoborder}\leaders\hrule height \headrulewidth\hfill}}
\renewcommand{\footrulewidth}{0.4pt}
\renewcommand{\footrule}{\hbox to\headwidth{\color{freedoborder}\leaders\hrule height \footrulewidth\hfill}}

\begin{document}
\thispagestyle{firstpage}

\vspace*{4pt}

\begin{center}
  {\LARGE\bfseries\color{freedodark} Zhengrong: World Models with Pole-Based\\Action-Response Factorization for Direct Visual Planning}\\[10pt]

  {\normalsize\bfseries\color{darkgray}
    Yu Zhang\textsuperscript{1},\quad
    Jingrui Li\textsuperscript{1},\quad
    Congyu Wang\textsuperscript{1},\quad
    Yuhang Cheng\textsuperscript{1},\quad
    Wenbo Chen\textsuperscript{1},\quad
    Xiaobo Zhang\textsuperscript{1},\quad
    Jing Zou\textsuperscript{1},\quad
    Yuxi Dai\textsuperscript{1}
  }\\[5pt]
  {\small\color{mutedgray}
    \textsuperscript{1}Beijing Freedo Technology Co., Ltd., Beijing, China
  }
\end{center}

\vspace{4pt}

\noindent
\begin{tikzpicture}
  \node[
    fill=freedolight,
    draw=freedoborder,
    line width=0.7pt,
    rounded corners=3pt,
    inner sep=11pt,
    text width=\dimexpr\textwidth-24pt\relax
  ] (absbox) {
    {\normalsize\bfseries\color{freedodark} Abstract}\\[4pt]
    {\small\color{darkgray}
    Visual model predictive control requires predicting how candidate action sequences unfold over future horizons. Direct world models eliminate recursive latent rollout, yet direct prediction interfaces leave the temporal organization of action effects unspecified. In this technical report, we introduce \textbf{Zhengrong} (\texttt{Z}hengrong \texttt{Pole}-based Action-Response \texttt{W}orld \texttt{M}odel, dubbed as \ours{}), which decomposes a latent trajectory into an initial-condition anchor, a free response, and causally delayed responses triggered by action innovations. Specifically, nonlinear networks map visual history and causal action prefixes to response coefficients, while a shared bank of stable conjugate poles governs their continuous evolution with elapsed time. This factorization preserves high-order nonlinear interactions across action sequences while guaranteeing exact initial-state anchoring, action-prefix non-anticipation, and bounded responses to finite control plans. Furthermore, we derive an exact re-anchoring identity that identifies the missing basis direction $\Psi_\kappa(s)$ required to represent shifted trajectory tails, formally separating single-response boundedness from restart closure. Across four offline image-goal control benchmarks (Cube, TwoRoom, PushT, and Reacher), \ours{} achieves $90.83\% \pm 0.47\%$ macro-average success over two training seeds at fixed epoch 20 and $91.21\% \pm 1.00\%$ under validation-loss checkpoint selection, outperforming the strongest baseline mean by $6.08$ percentage points. In a controlled first-epoch Cube study without restart regularization, neural and pole response heads improve over a generic direct predictor by $20.67$ and $23.34$ percentage points, respectively. On an NVIDIA A100 GPU, the optimized planner completes a full 30-iteration cross-entropy method search in $74\text{--}87\,\mathrm{ms}$ (mean $77.5\,\mathrm{ms}$), reducing latency by $15.8\%$ relative to Fast-LeWM and achieving a $7.1\times$ speedup over autoregressive LeWM rollout.}
  };
  \fill[freedoblue,rounded corners=1.5pt] (absbox.north west) rectangle ([xshift=3.5pt]absbox.south west);
\end{tikzpicture}

\vspace{6pt}

\section{Introduction}\label{sec:intro}

Visual world models enable goal-directed control by predicting the consequences of candidate actions in a learned representation space \citep{watter2015e2c,hafner2019planet,hafner2020dreamer}. In reward-free image-goal planning, joint-embedding predictive architectures (JEPA) offer a compact planning interface: the model encodes the current visual history and a target goal image, predicts the future embedding of a candidate action sequence, and minimizes the terminal distance to the goal representation via sampling-based optimization \citep{zhou2025dinowm,maes2026lewm}. Because an online planner such as the cross-entropy method (CEM) must evaluate hundreds of candidate plans over dozens of iterative updates at every decision step, the internal parameterization of the predictive dynamics governs both sample efficiency across prediction horizons and real-time planning latency.

Existing visual world models construct multi-step predictions through two primary paradigms. Autoregressive one-step predictors recursively apply a single-step transition operator to previously predicted states, as exemplified by LeWM \citep{maes2026lewm}. While recursive rollout naturally enforces Markovian step composition, it accumulates prediction errors over long horizons and incurs high sequential latency during iterative planning ($\approx 550\,\mathrm{ms}$ per 30-iteration CEM solve on an A100 GPU). Direct action-prefix predictors instead map the observed state and a causal action prefix directly to a future state representation in a single step, as demonstrated by Fast-LeWM \citep{gao2026fast}. Although direct prefix prediction removes recursive state rollout, it treats temporal querying as an unstructured black-box mapping and leaves a fundamental question unresolved: \emph{how should action effects share temporal structure across query times within a direct visual predictor?}

Consider a candidate control sequence that alters actuation at two distinct times. The first action change at $\tau_0$ initiates a dynamical response; the second change at $\tau_1$ initiates another response conditioned on the preceding control history. At any query time $t$, only responses whose onset precedes $t$ ($\tau_k \le t$) can contribute to the latent state. Generic direct predictors rely on unstructured neural layers to approximate this temporal coupling from training windows alone. In contrast, structuring the predictor as a superposition of an initial condition, a free response, and delayed action-innovation responses decouples \emph{what response an action initiates} from \emph{how that response unfolds over elapsed time}.

\begin{figure}[t!]
\centering
\begin{tikzpicture}[>=Latex, every node/.style={font=\small},
 box/.style={draw=freedodark!75,rounded corners=3pt,align=center,minimum height=.80cm},
 every path/.style={line width=.8pt,color=freedodark!85}]
\node[anchor=west,font=\small\bfseries,color=freedodark] at (0,3.6) {\ours{}: nonlinear response content + analytic temporal evolution};
\node[box,fill=poleblue!10,text width=2.5cm] (history) at (1.3,2.25) {Visual history\\state $\mathbf{q}_0$};
\node[box,fill=eventorange!10,text width=2.5cm] (actions) at (1.3,.7) {Candidate actions\\$\mathbf{u}_0,\ldots,\mathbf{u}_{K-1}$};
\node[box,fill=eventorange!10,text width=2.65cm,minimum height=1.08cm] (prefix) at (4.65,.7) {Causal prefix GRU\\innovation gate\\coefficients $\mathbf{c}_k$};
\node[box,fill=poleblue!10,text width=2.65cm] (free) at (4.65,2.25) {Free-response head\\$\mathbf{c}_{\mathsf{f}}(\mathbf{q}_0)$};
\node[box,fill=eventgreen!10,text width=3.1cm,minimum height=2.18cm] (bank) at (8.35,1.5) {
\textbf{Shared pole basis}\\[4pt]
$\mathbf{B}(t)\mathbf{c}_{\mathsf{f}}$\\[3pt]
$\mathbf{B}(t-\tau_0)\mathbf{c}_0$\\[3pt]
$\mathbf{B}(t-\tau_1)\mathbf{c}_1,\ldots$};
\node[box,fill=freedolight,text width=2.05cm,minimum height=1.42cm] (read) at (11.9,1.5) {Sum with $\mathbf{q}_0$\\readout $D_\theta$\\goal cost};
\draw[->] (history)--(free);
\draw[->] (actions)--(prefix);
\draw[->] (history.south east)--(prefix.north west);
\draw[->] (free.east)--(bank.north west);
\draw[->] (prefix.east)--(bank.south west);
\draw[->] (bank)--(read);
\draw[->] (history.north)--(1.3,3.02)--node[above,font=\scriptsize\bfseries,color=freedoblue] {initial-state anchor ($\widehat{\mathbf{q}}(0)=\mathbf{q}_0$)} (11.9,3.02)--(read.north);
\draw[freedoborder,line width=1pt] (0,-.10)--(13,-.10);
\node[anchor=west,font=\scriptsize,color=darkgray] at (0,-.44) {Query $t_1$: evaluate active responses at $t_1-\tau_k$};
\node[anchor=west,font=\scriptsize,color=darkgray] at (6.70,-.44) {Query $t_2$: reuse coefficients, change elapsed times};
\end{tikzpicture}
\caption{Architecture of \ours{} separating response content generation from temporal evolution. Visual history determines the free response $\mathbf{c}_{\mathsf{f}}(\mathbf{q}_0)$; causal action prefixes and bias-free innovation gates determine event responses $\mathbf{c}_k(\mathbf{q}_0,\mathbf{u}_{0:k})$. The shared pole basis evaluates active responses at elapsed times $t-\tau_k$ with zero contribution before onset.}
\label{fig:architecture}
\end{figure}
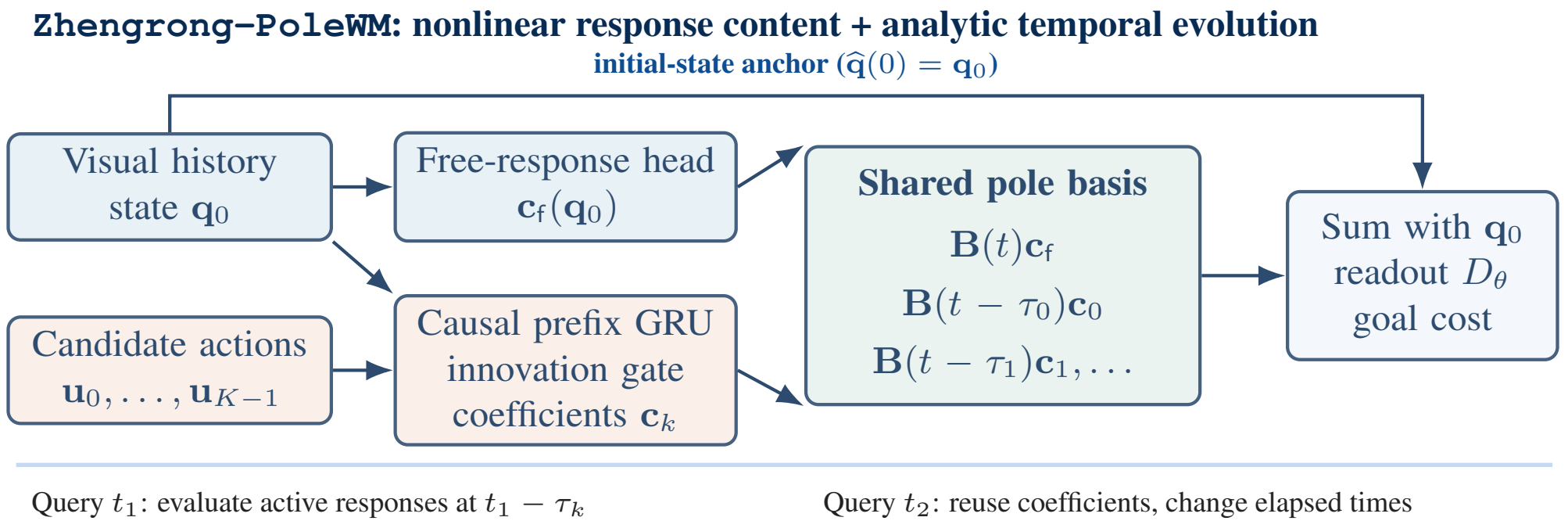

To operationalize this insight, we propose the \textbf{Zhengrong} world model (\texttt{Z}hengrong \texttt{Pole}-based Action-Response \texttt{W}orld \texttt{M}odel, dubbed as \ours{}), illustrated in Figure~\ref{fig:architecture}. \ours{} compresses observed visual frames and executed controls into an initial latent state $\mathbf{q}_0$, generates free-response coefficients from $\mathbf{q}_0$, and employs a causal action-prefix GRU coupled with a bias-free action-innovation gate $\Delta \mathbf{u}_k = \mathbf{u}_k - \mathbf{u}_{k-1}$ to generate event-specific drift and residue coefficients. A shared analytic temporal basis—comprising a real initial-value anchor $A(t)$ and second-order compensated complex exponentials $C_{p_m}(t)$ parameterized by 25 conjugate pole pairs in the left half-plane—evaluates every active response at elapsed time $t - \tau_k$. The second-order compensation ensures $A(0)=0, A'(0)=1$ and $C_{p_m}(0)=C'_{p_m}(0)=0$, orthogonally decoupling the onset velocity vector $\mathbf{v}_k$ from the transient curvature residues $\mathbf{b}_{k,m}$ without requiring numerical ODE integration or numerical inverse Laplace transformation (ILT). Furthermore, we prove that re-anchoring a compensated pole trajectory introduces an exponential-polynomial direction $\Psi_\kappa(s) = A(s) - s e^{-\kappa s}$ outside the original real span, rigorously explaining why single-trajectory boundedness and restart consistency represent distinct algebraic properties in direct predictors.

We evaluate \ours{} across three empirical levels and deliver a standalone, bitwise-verified software package (\texttt{pole\_worldmodel}, 18.13M parameters, 244 state tensors). Our contributions are three-fold:
\begin{itemize}[leftmargin=*]
  \item We propose \ours{}, a direct visual world model that factors latent trajectories into an initial-state anchor, a free response, and causally delayed action-innovation responses evaluated over a shared temporal basis (\S\ref{sec:method}).
  \item We construct an analytically compensated conjugate-pole basis with bias-free innovation gating, prove exact initial-state anchoring, action-prefix non-anticipation, and finite-plan boundedness, and derive the exact re-anchoring identity that characterizes the missing restart-closure direction $\Psi_\kappa(s)$ (\S\ref{sec:structure}).
  \item Across four visual control benchmarks (Cube, TwoRoom, PushT, and Reacher) with 300 tasks per environment, \ours{} achieves $90.83\%$ macro-average success at fixed epoch 20 ($+6.08$ percentage points over the strongest baseline), improves early Cube control by $+23.34$ points over a generic direct head, and completes a 30-iteration CEM planning solve in $74\text{--}87\,\mathrm{ms}$ on an A100 GPU (\S\ref{sec:experiments}).
\end{itemize}

\section{Related work}\label{sec:related}

\begin{table}[t!]
\centering
\caption{Conceptual comparison between \ours{} and prior world-model paradigms across five structural dimensions.}
\label{tab:conceptual}
\fontsize{8}{10}\selectfont
\setlength{\tabcolsep}{0.88pt}
\begin{tabular}{lccccc}
\toprule
Paradigm & Direct Query & Shared Basis & Event Anchor & Analytic Eval. & Prefix Nonlinearity \\
\midrule
Autoregressive MBRL \citep{hafner2019planet,maes2026lewm} & No (Rollout) & No & Markov Step & N/A & Yes (Recurrent) \\
Pretrained-Feature WM \citep{zhou2025dinowm} & No (Rollout) & No & Markov Step & N/A & Yes (Transformer) \\
Neural Laplace \citep{holt2022laplace,holt2023control} & Yes & Laplace Domain & Delayed $s$-Rep. & No (Num.\ ILT) & Yes \\
Generic Direct WM \citep{gao2026fast} & Yes & No (MLP) & Approximate & Yes (Forward) & Yes \\
\rowcolor{tabhighlight}
\ours{} (ours) & Yes & Yes (Poles) & Exact ($\widehat{\mathbf{q}}(0)=\mathbf{q}_0$) & Yes (Closed-Form) & Yes (GRU + Gate) \\
\bottomrule
\end{tabular}
\end{table}

\noindent\textbf{Latent dynamics and visual control.}
Learning control-oriented latent representations from high-dimensional images has been a foundational paradigm in model-based control. E2C learns locally linear latent transitions from raw pixels \citep{watter2015e2c}, PETS combines uncertainty-aware ensembles with sampling-based CEM planning \citep{chua2018pets}, and PlaNet couples recurrent state-space models (RSSM) with online model predictive control \citep{hafner2019planet}. Dreamer learns policies through latent imagination \citep{hafner2020dreamer}, while TD-MPC and TD-MPC2 integrate latent trajectory optimization with learned terminal value functions \citep{hansen2022tdmpc,hansen2024tdmpc2}. While these systems rely on reward supervision or online policy learning, \ours{} addresses reward-free, offline image-goal planning driven solely by terminal embedding distances.

\noindent\textbf{Predictive visual representations and direct world models.}
Self-supervised visual representation learning avoids pixel reconstruction to capture semantic and geometric structure: DINO uses self-distillation \citep{caron2021dino}, I-JEPA predicts target representations from context blocks \citep{assran2023ijepa}, and VICReg prevents representation collapse via variance-invariance-covariance regularization \citep{bardes2022vicreg}. For visual planning, DINO-WM plans over pretrained DINO features \citep{zhou2025dinowm}, LeWM jointly trains visual embeddings and one-step latent dynamics \citep{maes2026lewm}, and QQWorld regularizes latent embeddings toward a Gaussian prior via sliced quantile-quantile matching \citep{yu2026qq}. Fast-LeWM replaces recursive one-step rollout with direct action-prefix prediction \citep{gao2026fast}. \ours{} adopts end-to-end visual representation learning and sliced QQ regularization while structuring the direct prediction head as a sum of free and causally delayed action-innovation responses (Table~\ref{tab:conceptual}).

\noindent\textbf{Response models and structured state spaces.}
Decomposing system trajectories into an initial-condition free response and a forced input response is a cornerstone of classical control theory \citep{astrom2008feedback}. In sequence modeling, S4 introduces structured state-space kernels \citep{gu2022s4}, and Mamba incorporates input-dependent selection mechanisms \citep{gu2024mamba}. \ours{} bridges classical response superposition with nonlinear prefix conditioning: each action innovation gates a prefix-conditioned coefficient vector, and its contribution is evaluated analytically at the time elapsed since the event onset.

\noindent\textbf{Operator learning and continuous-time prediction.}
DeepONet factors operator learning into input-function encoding (branch network) and output-coordinate evaluation (trunk network) \citep{lu2021deeponet}, while Fourier neural operators (FNO) parameterize integral kernels in frequency space \citep{li2021fno}. \ours{} shares the coefficient--coordinate factorization perspective while enforcing exact initial-state anchoring, causal action-prefix conditioning, and event-wise temporal shifts. Unlike Neural ODEs \citep{chen2018node}, which integrate vector fields numerically, or Neural Laplace \citep{holt2022laplace,holt2023control} and Laplace Neural Operators (LNO) \citep{cao2024lno}, which operate in the transform domain or rely on numerical inverse Laplace transformation, \ours{} evaluates its second-order compensated pole basis directly in the time domain in closed form.

\noindent\textbf{Spectral dynamics and composition consistency.}
Dynamic mode decomposition with control (DMDc) separates autonomous evolution from linear actuation \citep{proctor2016dmdc}, and deep Koopman models seek globally linearizing coordinate embeddings \citep{lusch2018koopman}. In \ours{}, the conjugate poles parameterize latent response waveforms rather than physical eigenmodes of a global linear operator, as history compression, action-prefix conditioning, and visual readout remain nonlinear. Complementing recent latent composition regularizers such as BRo-JEPA \citep{jha2026bro} and group-action world models \citep{wang2026group}, our analysis in \S\ref{sec:structure} establishes the algebraic conditions separating single-trajectory pole boundedness from exact re-anchoring closure.

\section{Preliminaries and problem formulation}\label{sec:prelim}

We first formalize the offline visual trajectory dataset, the multi-scale model clock, and the direct action-prefix planning interface used throughout this work.

\noindent\textbf{Offline visual trajectory data and history alignment.}
We learn from an offline corpus of interaction episodes without reward annotations. Each training window consists of $J = 8$ strided RGB observations $\mathbf{o}_i \in \R^{3\times H\times W}$ ($H = W = 224$) separated by macro action blocks. The first $L = 3$ frames form the observation history at indices $i \in \{-2, -1, 0\}$, and the subsequent $5$ frames at macro steps $j \in \{1, 2, 3, 4, 5\}$ provide future prediction targets at model times $t_j = j\Delta t$ with macro interval $\Delta t = 0.25$. Each macro action block $\mathbf{U}_i \in \R^{b \cdot d_u}$ groups $b = 5$ consecutive normalized raw controls of dimension $d_u$, corresponding to a micro-event interval $\delta = \Delta t / b = 0.05$.

An end-to-end visual encoder and projector $E_\theta$ map each RGB observation $\mathbf{o}_i$ to a $D$-dimensional visual embedding $\mathbf{e}_i = E_\theta(\mathbf{o}_i) \in \R^D$ ($D = 192$). To prevent future action leakage during history aggregation, the three history frames and two intervening executed macro action blocks are aligned into three causal history slots as follows:
\begin{equation}\label{eq:history_slots}
\mathcal{H}_0 = \bigl[(\mathbf{e}_{-2}, \mathbf{U}_{-2}),\; (\mathbf{e}_{-1}, \mathbf{U}_{-1}),\; (\mathbf{e}_0, \mathbf{0})\bigr],
\end{equation}
where each nonzero action slot carries the outgoing controls leading to the next observed frame, and the current frame's outgoing slot is zero-padded. A recurrent history encoder $H_\theta$ (a GRU with a residual connection to the latest visual embedding $\mathbf{e}_0$) compresses $\mathcal{H}_0$ into a single initial latent state, and a residual MLP readout $D_\theta$ maps predicted latent states back to the visual embedding space:
\begin{equation}\label{eq:state}
\mathbf{q}_0 = H_\theta\bigl(\mathbf{e}_{-L+1:0},\, \mathbf{U}_{-L+1:-1}\bigr) \in \R^D, \qquad
\widehat{\mathbf{e}}(t) = D_\theta\bigl(\widehat{\mathbf{q}}(t)\bigr) \in \R^D,
\end{equation}
where $\mathbf{q}_0$ summarizes history-dependent dynamical context and $\widehat{\mathbf{e}}(t)$ lives in the visual representation space used to evaluate terminal image-goal costs.

\noindent\textbf{Direct action-prefix prediction Problem.}
Given an initial history state $\mathbf{q}_0$ and a candidate sequence of $K$ normalized raw controls $\mathbf{u} = (\mathbf{u}_0, \mathbf{u}_1, \ldots, \mathbf{u}_{K-1})$ occurring at absolute event onset times $\tau_k = k\delta$ ($k = 0, \ldots, K-1$), a direct world model defines a continuous-time trajectory predictor $F_\theta(t; \mathbf{q}_0, \mathbf{u}) = \widehat{\mathbf{q}}(t)$ for any query time $t \ge 0$. A generic unstructured direct head parameterizes $F_\theta$ as a black-box neural mapping from $(\mathbf{q}_0, \mathbf{u}_{0:k}, t)$ to $\widehat{\mathbf{q}}(t)$, which neither guarantees initial-condition consistency $\widehat{\mathbf{q}}(0) = \mathbf{q}_0$ nor shares temporal evolution structure across distinct query times $t_1 \neq t_2$.

\section{Methodology: Zhengrong action-response world model}\label{sec:method}

We now present \ours{}. We first introduce the action-response trajectory factorization (\S\ref{sec:factorization}). After that, we construct the second-order compensated conjugate-pole basis and bias-free innovation gate (\S\ref{sec:pole_basis}). Then, we derive the equivalent Laplace-domain representation and numerical stabilization scheme (\S\ref{sec:laplace_rep}). Finally, we formulate the joint representation-dynamics training objective and CEM planning loop (\S\ref{sec:learning_planning}).

\subsection{Action-response trajectory factorization}\label{sec:factorization}

To share temporal structure across query times while preserving nonlinear action conditioning, \ours{} factors the latent trajectory $\widehat{\mathbf{q}}(t)$ into the initial state $\mathbf{q}_0$, a free response driven by $\mathbf{q}_0$, and a causal sum of delayed event responses triggered at onset times $\tau_k = k\delta$:
\begin{equation}\label{eq:factorization}
\widehat{\mathbf{q}}(t) = \mathbf{q}_0 + \mathbf{B}(t)\mathbf{c}_{\mathsf{f}}(\mathbf{q}_0)
 + \sum_{k:\,\tau_k \le t} \mathbf{B}(t - \tau_k)\,\mathbf{c}_k(\mathbf{q}_0, \mathbf{u}_{0:k}), \qquad \mathbf{B}(0) = \mathbf{0},
\end{equation}
where $\mathbf{B}(t)$ denotes a shared bank of scalar temporal basis functions and the product contracts basis channels with vector-valued coefficients $\mathbf{c}_{\mathsf{f}}, \mathbf{c}_k$. Intuitively, the free response $\mathbf{B}(t)\mathbf{c}_{\mathsf{f}}(\mathbf{q}_0)$ captures autonomous evolution from the observed history state, while the $k$-th term $\mathbf{B}(t - \tau_k)\mathbf{c}_k(\mathbf{q}_0, \mathbf{u}_{0:k})$ captures the delayed dynamical response initiated at $\tau_k$.

\noindent\textbf{Inductive bias of response factorization.}
Equation~\eqref{eq:factorization} ties predictions at arbitrary query horizons through a shared temporal response family. A later control input $\mathbf{u}_k$ modulates its own response coefficient $\mathbf{c}_k(\mathbf{q}_0, \mathbf{u}_{0:k})$ through causal prefix conditioning, yet cannot alter any response contribution prior to its onset $\tau_k$. This design cleanly separates three functional roles: (i) the observed visual history sets the initial state $\mathbf{q}_0$; (ii) the causal action prefix determines the nonlinear response content $\mathbf{c}_k$; and (iii) the elapsed time $t - \tau_k$ evaluates each active waveform on the shared temporal basis. Because the same coefficients $\{\mathbf{c}_{\mathsf{f}}, \mathbf{c}_0, \ldots, \mathbf{c}_{K-1}\}$ serve all query times $t$ for a given history and plan, multi-horizon supervision directly constrains a single shared trajectory.

\subsection{Analytic conjugate-pole basis and bias-free action innovations}\label{sec:pole_basis}

We instantiate the shared temporal basis $\mathbf{B}(t)$ using a real anchor function and $M$ pairs of second-order compensated complex exponentials. For a fixed anchor decay rate $\kappa > 0$ ($\kappa = 1.0$) and complex pole $p \in \C$, we define the scalar basis functions for $t \ge 0$ as:
\begin{equation}\label{eq:basis}
A(t) = \frac{1 - e^{-\kappa t}}{\kappa}, \qquad
C_p(t) = e^{pt} - e^{-\kappa t} - (p + \kappa)t e^{-\kappa t},
\end{equation}
where $A(t)$ and $C_p(t)$ satisfy the onset boundary conditions $A(0) = 0$, $A'(0) = 1$, and $C_p(0) = C'_p(0) = 0$. The two subtraction terms $-e^{-\kappa t} - (p+\kappa)te^{-\kappa t}$ in $C_p(t)$ algebraically remove both the initial value and initial first derivative of the oscillatory exponential $e^{pt}$ at $t=0$.

With $M = 25$ conjugate pole pairs (50 complex poles in total), we parameterize the upper-half-plane representative $p_m = -d_m + \mathrm{i}\omega_m$ of each conjugate pair $(p_m, \overline{p_m})$ via strictly positive decay and frequency floors:
\begin{equation}\label{eq:poles}
p_m = -\bigl[d_{\min} + \spplus(\alpha_m)\bigr]
 + \mathrm{i}\bigl[\omega_{\min} + \spplus(\beta_m)\bigr],
\qquad d_{\min} = 0.05,\;\; \omega_{\min} = 0.1,
\end{equation}
where $\alpha_m, \beta_m \in \R$ are learnable pole parameters. Enforcing $\operatorname{Re}p_m \le -d_{\min} < 0$ ensures that every oscillatory mode $C_{p_m}(t)$ decays exponentially over elapsed time, while the real anchor $A(t) \to \kappa^{-1}$ as $t \to \infty$ captures persistent displacement in latent state space.

For a real drift vector $\mathbf{v} \in \R^D$ and complex residue vectors $\mathbf{b} = \{\mathbf{b}_m\}_{m=1}^M$ with $\mathbf{b}_m \in \C^D$, we combine each conjugate pair in the time domain to form a real-valued response operator:
\begin{equation}\label{eq:phi}
\boldsymbol{\phi}(t;\, \mathbf{v}, \mathbf{b}) = A(t)\mathbf{v} + 2\operatorname{Re}\sum_{m=1}^{M} C_{p_m}(t)\mathbf{b}_m,
\end{equation}
where conjugate pairing guarantees that $\boldsymbol{\phi}(t; \mathbf{v}, \mathbf{b}) \in \R^D$ for all $t \ge 0$. Because $A'(0) = 1$ and $C'_{p_m}(0) = 0$, the coefficient groups enjoy orthogonal kinematic roles at onset: the drift vector $\mathbf{v} = \left.\frac{\mathrm{d}}{\mathrm{d}t}\boldsymbol{\phi}(t; \mathbf{v}, \mathbf{b})\right|_{t=0^+}$ exclusively governs the initial onset velocity (slope), whereas the complex residues $\mathbf{b}_m$ shape higher-order transient curvature without perturbing the initial value or initial slope.

\noindent\textbf{Causal prefix conditioning and bias-free innovation gate.}
A free-response MLP head maps $\mathbf{q}_0$ directly to the free coefficients $(\mathbf{v}_{\mathsf{f}}, \mathbf{b}_{\mathsf{f}})$. To generate action-event coefficients, a unidirectional action-prefix GRU initialized from $\mathbf{q}_0$ sequentially processes the tuple $(\mathbf{u}_k, \tau_k)$ to update its hidden state $\mathbf{h}_k \in \R^{D_{\mathsf{dyn}}}$ ($D_{\mathsf{dyn}} = 512$). We generate the $k$-th event coefficient vector $\boldsymbol{\eta}_k \in \R^{(1+2M)D}$ through a multiplicative bias-free innovation gate:
\begin{equation}\label{eq:gate}
\boldsymbol{\eta}_k = \mathbf{g}_\theta\bigl([\mathbf{q}_0, \mathbf{h}_k]\bigr) \odot \mathbf{W}_\theta(\mathbf{u}_k - \mathbf{u}_{k-1}), \qquad
(\mathbf{v}_k, \mathbf{b}_k) = \operatorname{unpack}(\boldsymbol{\eta}_k),
\end{equation}
where $\mathbf{g}_\theta$ is a nonlinear conditioning MLP, $\odot$ denotes the element-wise Hadamard product, and $\mathbf{W}_\theta \in \R^{(1+2M)D \times d_u}$ is a linear projection matrix \emph{without bias}. In $\operatorname{unpack}(\boldsymbol{\eta}_k)$, the $(1+2M)D$ real channels partition into $D$ real drift coordinates $\mathbf{v}_k$ and $2MD$ real and imaginary residue coordinates $\mathbf{b}_k$. Because $\mathbf{W}_\theta$ has no bias term, maintaining a constant action ($\mathbf{u}_k = \mathbf{u}_{k-1}$) yields $\mathbf{W}_\theta(\mathbf{u}_k - \mathbf{u}_{k-1}) = \mathbf{0}$ and thus $\boldsymbol{\eta}_k = \mathbf{0}$; an unchanged control triggers no spurious impulse response while earlier responses continue to evolve naturally.

We set $\mathbf{u}_{-1} = \mathbf{0}$ prior to the plan and append a terminal action $\mathbf{u}_K = \mathbf{0}$ at $\tau_K = K\delta$ to specify continuation after the $K$-step finite horizon in normalized action coordinates. Assembling the free response and all $K+1$ causal event responses yields the complete \ours{} trajectory predictor:
\begin{equation}\label{eq:response}
\widehat{\mathbf{q}}(t) = \mathbf{q}_0 + \boldsymbol{\phi}(t;\, \mathbf{v}_{\mathsf{f}}, \mathbf{b}_{\mathsf{f}})
 + \sum_{k=0}^{K} \mathbf{1}_{t \ge \tau_k}\,\boldsymbol{\phi}\bigl(t - \tau_k;\, \mathbf{v}_k, \mathbf{b}_k\bigr),
\end{equation}
where $\mathbf{1}_{t \ge \tau_k}$ is the Heaviside step indicator. At the planning horizon endpoint $t = K\delta = \tau_K$, the terminal event $k = K$ has zero elapsed time ($t - \tau_K = 0$) and contributes $\boldsymbol{\phi}(0; \mathbf{v}_K, \mathbf{b}_K) = \mathbf{0}$, leaving endpoint predictions unchanged.

\subsection{Equivalent Laplace representation and numerical stabilization}\label{sec:laplace_rep}

Although \ours{} evaluates trajectories analytically in the time domain, examining its exact Laplace transform clarifies the frequency-domain structure being learned. For $\operatorname{Re}s > 0$, the Laplace transforms of the anchor $A(t)$ and compensated pole basis $C_p(t)$ take the rational forms:
\begin{equation}\label{eq:laplace_basis}
\widetilde{A}(s) = \mathcal{L}\{A\}(s) = \frac{1}{s(s+\kappa)}, \qquad
G_p(s) = \mathcal{L}\{C_p\}(s) = \frac{(p+\kappa)^2}{(s-p)(s+\kappa)^2},
\end{equation}
where the algebraic compensation in $C_p(t)$ produces a triple pole denominator $(s-p)(s+\kappa)^2$ with $\mathcal{O}(|s|^{-3})$ high-frequency roll-off. Defining the conjugate-paired transform operator $\mathcal{B}(s; \mathbf{b}) = \sum_{m=1}^M \bigl[G_{p_m}(s)\mathbf{b}_m + G_{\overline{p_m}}(s)\overline{\mathbf{b}_m}\bigr]$, the Laplace transform $\widehat{\mathbf{Q}}(s) = \mathcal{L}\{\widehat{\mathbf{q}}\}(s)$ of the full trajectory in Eq.~\eqref{eq:response} is:
\begin{equation}\label{eq:full_laplace}
\widehat{\mathbf{Q}}(s) = \frac{\mathbf{q}_0}{s} + \widetilde{A}(s)\mathbf{v}_{\mathsf{f}} + \mathcal{B}(s;\, \mathbf{b}_{\mathsf{f}})
 + \sum_{k=0}^{K} e^{-s\tau_k}\bigl[\widetilde{A}(s)\mathbf{v}_k + \mathcal{B}(s;\, \mathbf{b}_k)\bigr],
\end{equation}
where conjugation in $\mathcal{B}(s; \mathbf{b})$ applies strictly to the poles $p_m$ and coefficients $\mathbf{b}_m$ rather than the complex frequency variable $s$. Equation~\eqref{eq:full_laplace} shows that each undelayed response is rational in $s$, whereas causal action innovations enter as exponential time-delay factors $e^{-s\tau_k}$.

\noindent\textbf{Near-zero Taylor stabilization and mixed-precision accumulation.}
Because Eq.~\eqref{eq:basis} is already the exact analytical inverse Laplace transform of Eq.~\eqref{eq:laplace_basis}, \ours{} completely avoids numerical inverse Laplace transform (ILT) quadrature during training and planning. In floating-point evaluation, directly computing $C_p(t) = e^{pt} - e^{-\kappa t} - (p+\kappa)te^{-\kappa t}$ when $w = (p+\kappa)t$ is near zero can suffer from catastrophic subtractive cancellation. To guarantee machine-precision accuracy near event onsets, our implementation factors $C_p(t) = e^{-\kappa t}\bigl(e^w - 1 - w\bigr)$ and switches to a fourth-order Taylor series expansion whenever $|w| < 10^{-3}$:
\begin{equation}\label{eq:taylor_guard}
C_p(t) = e^{-\kappa t}\left(\frac{w^2}{2} + \frac{w^3}{6} + \frac{w^4}{24} + \frac{w^5}{120} + \mathcal{O}(w^6)\right), \qquad w = (p + \kappa)t.
\end{equation}
All neural coefficient networks operate in FP32, while basis evaluation in Eq.~\eqref{eq:basis}--\eqref{eq:taylor_guard} and cross-event summation in Eq.~\eqref{eq:response} are executed in FP64 / \texttt{complex128} before casting the resulting state trajectory $\widehat{\mathbf{q}}(t)$ back to FP32.

\subsection{Joint representation-dynamics learning and CEM planning}\label{sec:learning_planning}

\noindent\textbf{Multi-time prediction, state, and readout losses.}
For each training window, sliding the history encoder $H_\theta$ across the five future macro steps $t_j \in \{0.25, 0.50, 0.75, 1.00, 1.25\}$ yields target visual embeddings $\mathbf{e}_j = E_\theta(\mathbf{o}_j)$ and target history states $\mathbf{q}_j$. We supervise the predicted trajectory $\widehat{\mathbf{q}}(t_j) = F_\theta(t_j; \mathbf{q}_0, \mathbf{u})$ with three mean-squared error terms averaged over batch size, target horizons, and feature dimensions:
\begin{equation}\label{eq:predictionloss}
\mathcal{L}_{\mathsf{pred}} = \MSE\bigl(D_\theta(\widehat{\mathbf{q}}(t_j)),\, \mathbf{e}_j\bigr), \quad
\mathcal{L}_{\mathsf{state}} = \MSE\bigl(\widehat{\mathbf{q}}(t_j),\, \mathbf{q}_j\bigr), \quad
\mathcal{L}_{\mathsf{read}} = \MSE\bigl(D_\theta(\mathbf{q}_j),\, \mathbf{e}_j\bigr),
\end{equation}
where $\mathcal{L}_{\mathsf{pred}}$ aligns decoded predictions with future visual embeddings, $\mathcal{L}_{\mathsf{state}}$ aligns predicted states with sliding history states, and $\mathcal{L}_{\mathsf{read}}$ ensures readout consistency on observed states. Both target and prediction branches receive full gradients without an exponential moving average (EMA) target encoder.

\noindent\textbf{Sliced Quantile-Quantile (QQ) embedding regularization.}
To prevent representation collapse in end-to-end training without an EMA teacher, we apply sliced quantile matching \citep{yu2026qq} to the observed visual embeddings $\mathbf{e}_j$ across all $J = 8$ frame positions of the training window. Given $P = 1024$ random unit projection directions $\{\mathbf{r}_\ell\}_{\ell=1}^P$ in $\R^D$, let $z_{j,(i),\ell}$ denote the $i$-th order statistic (sorted projection) across the global batch of $B = 128$ samples at frame position $j$:
\begin{equation}\label{eq:qq_loss}
\mathcal{L}_{\mathsf{QQ}} = \frac{1}{JPB}\sum_{j=1}^{J}\sum_{\ell=1}^{P}\sum_{i=1}^{B}
\left[z_{j,(i),\ell} - \Phi^{-1}\!\left(\frac{i - 1/2}{B}\right)\right]^2,
\end{equation}
where $\Phi^{-1}$ is the standard normal inverse cumulative distribution function. Under distributed data-parallel (DDP) training, embeddings are gathered differentiably across ranks and projection directions are sharded across devices in FP32.

\noindent\textbf{Auxiliary restart flow consistency and total objective.}
To encourage multi-step composition consistency when a trajectory is re-initialized from an intermediate predicted state $\widehat{\mathbf{q}}(t_r)$ at macro boundary $r \in \{1, 2, 3, 4\}$ with suffix control sequence $\mathbf{u}^{(r)}$, we define the auxiliary restart flow loss:
\begin{equation}\label{eq:flow}
\mathcal{L}_{\mathsf{flow}} = \E_r\,\MSE_{j > r}\!\left(
F_\theta\bigl(t_j - t_r;\, \widehat{\mathbf{q}}(t_r),\, \mathbf{u}^{(r)}\bigr),\; \sg\bigl[\widehat{\mathbf{q}}(t_j)\bigr]\right),
\end{equation}
where $\sg[\cdot]$ stops gradients on the direct tail while keeping the intermediate restart state $\widehat{\mathbf{q}}(t_r)$ differentiable. Each restart re-initializes the action-prefix GRU from $\widehat{\mathbf{q}}(t_r)$ and resets the preceding normalized action to zero. Combining Eq.~\eqref{eq:predictionloss}--\eqref{eq:flow}, the total training objective is:
\begin{equation}\label{eq:loss}
\mathcal{L} = \mathcal{L}_{\mathsf{pred}} + 0.25\mathcal{L}_{\mathsf{state}}
 + 0.25\mathcal{L}_{\mathsf{read}} + 3.5\mathcal{L}_{\mathsf{QQ}}
 + \lambda_{\mathsf{flow}}\mathcal{L}_{\mathsf{flow}},
\end{equation}
where $\lambda_{\mathsf{flow}} = 0.05$ for the full model and $\lambda_{\mathsf{flow}} = 0$ for all variants in the controlled early response-head study (\S\ref{sec:ablation}). To align training with single-frame cold-start control at episode reset, cold-history augmentation replaces the 3-frame history with repeated current embeddings $[\mathbf{e}_0, \mathbf{e}_0, \mathbf{e}_0]$ and zero past actions in $25\%$ of training windows.

\noindent\textbf{Goal-conditioned CEM planning.}
At planning time, given a goal observation $\mathbf{o}_g$ with target embedding $\mathbf{e}_g = E_\theta(\mathbf{o}_g)$, the CEM planner searches over candidate control sequences $\mathbf{u} = (\mathbf{u}_0, \ldots, \mathbf{u}_{K-1})$ ($K = 25$) to minimize the squared terminal distance in visual embedding space:
\begin{equation}\label{eq:cost}
J(\mathbf{u}) = \bigl\|D_\theta\bigl(F_\theta(K\delta;\, \mathbf{q}_0,\, \mathbf{u})\bigr) - \mathbf{e}_g\bigr\|_2^2,
\end{equation}
where $\mathbf{q}_0$, $\mathbf{e}_g$, and the free response $\boldsymbol{\phi}(K\delta; \mathbf{v}_{\mathsf{f}}, \mathbf{b}_{\mathsf{f}})$ depend only on the observed history and are cached once per CEM solve across all $300$ candidates and $30$ search iterations. Table~\ref{tab:algorithm} summarizes the end-to-end training and online planning loops.

\begin{table}[t!]
\centering
\caption{Summary of the end-to-end offline training procedure and online cached-embedding CEM planning loop.}
\label{tab:algorithm}
\fontsize{8}{10}\selectfont
\setlength{\tabcolsep}{6pt}
\begin{tabularx}{\linewidth}{@{}lX@{}}
\toprule
Phase & Execution Steps \\
\midrule
Offline Training & 1. Encode 8-frame window $\mathbf{o}_{-2:5}$ via $E_\theta$ to obtain $\mathbf{e}_{-2:5}$; apply cold-start augmentation with probability $0.25$. \\
 & 2. Construct initial history state $\mathbf{q}_0$ and sliding target states $\mathbf{q}_{1:5}$ via $H_\theta$ using causal alignment Eq.~\eqref{eq:history_slots}. \\
 & 3. Compute free response $(\mathbf{v}_{\mathsf{f}}, \mathbf{b}_{\mathsf{f}})$ and prefix-conditioned event coefficients $(\mathbf{v}_k, \mathbf{b}_k)$ via Eq.~\eqref{eq:gate}. \\
 & 4. Evaluate $\widehat{\mathbf{q}}(t_j)$ at $t_j \in \{0.25, \ldots, 1.25\}$ in FP64/\texttt{complex128} via Eq.~\eqref{eq:response}; compute $\mathcal{L}_{\mathsf{pred}}, \mathcal{L}_{\mathsf{state}}, \mathcal{L}_{\mathsf{read}}, \mathcal{L}_{\mathsf{QQ}}, \mathcal{L}_{\mathsf{flow}}$ and update $\theta$. \\
\midrule
Online CEM Planning & 1. Encode observed history and goal $\mathbf{o}_g$ once; cache initial state $\mathbf{q}_0$, goal embedding $\mathbf{e}_g$, and free response $\boldsymbol{\phi}(K\delta; \mathbf{v}_{\mathsf{f}}, \mathbf{b}_{\mathsf{f}})$. \\
 & 2. In each of $30$ CEM iterations, sample $300$ candidate plans $\mathbf{u}$ ($K=25$ raw controls), convert to model-normalized coordinates, and evaluate terminal states $\widehat{\mathbf{q}}(K\delta)$ and goal costs $J(\mathbf{u})$ in Eq.~\eqref{eq:cost}. \\
 & 3. Refit Gaussian sampling distribution to top $30$ elites; execute initial control block in the environment and replan with real history. \\
\bottomrule
\end{tabularx}
\end{table}

\section{Theoretical analysis, structural guarantees, and operator connections}\label{sec:structure}

We now analyze the mathematical properties guaranteed by the \ours{} parameterization. We first establish exact anchoring, action non-anticipation, and finite-plan boundedness (\S\ref{sec:props}). After that, we prove the exact re-anchoring identity that characterizes the missing restart-closure direction $\Psi_\kappa(s)$ (\S\ref{sec:restart_theory}). Finally, we formalize the conditional connection to Koopman generator resolvents and derive a finite-pool decision regret bound (\S\ref{sec:koopman_regret}). Complete proofs are provided in \S\ref{app:proofs}.

\subsection{Exact anchoring, non-anticipation, and finite-plan boundedness}\label{sec:props}

\begin{proposition}[Initial-state anchoring, continuity, and action non-anticipation]\label{prop:causal}
For any finite initial state $\mathbf{q}_0$, real drift coefficients, and conjugate-paired complex residues, the trajectory $\widehat{\mathbf{q}}(t)$ in Eq.~\eqref{eq:response} is real-valued and continuous on $[0, \infty)$ with $\widehat{\mathbf{q}}(0) = \mathbf{q}_0$. Any action event occurring at or after query time $t$ ($\tau_k \ge t$) has zero effect on $\widehat{\mathbf{q}}(t)$, and appending arbitrary normalized-zero controls preserves the entire predicted trajectory function identically.
\end{proposition}

Because $A(0) = C_{p_m}(0) = C'_{p_m}(0) = 0$ and $A'(0) = 1$, each event response vanishes at its onset $\tau_k$ and changes the right derivative by exactly $\mathbf{v}_k$ without creating a state jump. Combined with unidirectional prefix conditioning and the bias-free innovation gate $\mathbf{W}_\theta(\mathbf{u}_k - \mathbf{u}_{k-1})$, Proposition~\ref{prop:causal} guarantees causal information flow by construction rather than through learned weight cancellation.

\begin{proposition}[Bounded finite-plan response and asymptotic limit]\label{prop:bounded}
For any fixed finite initial state $\mathbf{q}_0$, finite action plan $\mathbf{u}_{0:K-1}$, finite response coefficients, $\kappa > 0$, and $\operatorname{Re}p_m < 0$, the predicted trajectory $\widehat{\mathbf{q}}(t)$ in Eq.~\eqref{eq:response} is uniformly bounded for all $t \ge 0$:
\begin{equation}\label{eq:explicit_bound}
\sup_{t \ge 0}\|\widehat{\mathbf{q}}(t)\|_2 \le \|\mathbf{q}_0\|_2 + \sum_{r \in \{\mathsf{f}, 0, \ldots, K\}}\left[\frac{\|\mathbf{v}_r\|_2}{\kappa} + 2\sum_{m=1}^{M}\left(2 + \frac{|p_m + \kappa|}{e\kappa}\right)\|\mathbf{b}_{r,m}\|_2\right] < \infty,
\end{equation}
and converges to the explicit steady-state limit:
\begin{equation}\label{eq:limit}
\lim_{t \to \infty}\widehat{\mathbf{q}}(t) = \mathbf{q}_0 + \kappa^{-1}\left(\mathbf{v}_{\mathsf{f}} + \sum_{k=0}^{K}\mathbf{v}_k\right),
\end{equation}
where $e = \exp(1)$ and each transient mode $C_{p_m}(t - \tau_k)$ decays exponentially to zero.
\end{proposition}

Proposition~\ref{prop:bounded} holds pointwise for any fixed state and finite control sequence, ensuring that open-loop multi-horizon queries never diverge at long evaluation times.

\subsection{Re-anchoring identity and restart closure analysis}\label{sec:restart_theory}

A natural question is whether exact initial anchoring and decaying pole transients automatically imply exact restart consistency (\ie whether the tail of a trajectory re-anchored at an intermediate time $r > 0$ with elapsed time $s = t - r \ge 0$ lies within the original response basis). Define the auxiliary exponential-polynomial function $\Psi_\kappa(s) = A(s) - s e^{-\kappa s} = \frac{1 - e^{-\kappa s}}{\kappa} - s e^{-\kappa s}$. Direct algebraic expansion establishes the following exact re-anchoring identity:
\begin{equation}\label{eq:closure}
C_p(r + s) - C_p(r) = e^{pr}C_p(s) + C'_p(r)A(s) + (p + \kappa)\bigl(e^{-\kappa r} - e^{pr}\bigr)\Psi_\kappa(s),
\end{equation}
where $C'_p(r) = p e^{pr} + \kappa e^{-\kappa r} - (p+\kappa)(1-\kappa r)e^{-\kappa r}$. For distinct nonreal poles $p_m$, the complex exponentials $\{e^{p_m s}, e^{\overline{p_m}s}\}$ are linearly independent of the anchor functions $\{1, e^{-\kappa s}, s e^{-\kappa s}\}$. Because $A(s)$ contains no $s e^{-\kappa s}$ term once all $C_{p_m}(s)$ coefficients are set to zero, \textbf{the direction $\Psi_\kappa(s)$ lies strictly outside the real linear span} of $\{A(s), \operatorname{Re}C_{p_m}(s), \allowbreak\operatorname{Im}C_{p_m}(s)\}_{m=1}^M$.

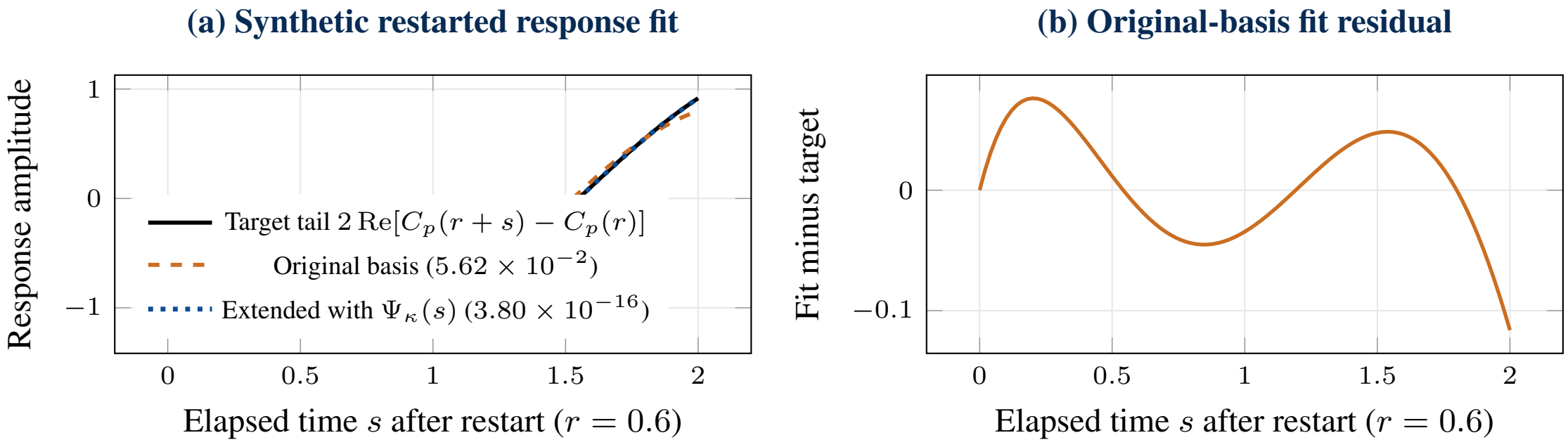
\begin{figure}[t!]
\centering
\begin{tikzpicture}
\begin{groupplot}[group style={group size=2 by 1,horizontal sep=1.65cm},
 width=.44\linewidth,height=4.2cm,xlabel={Elapsed time $s$ after restart ($r=0.6$)},
 tick label style={font=\scriptsize},label style={font=\small},
 title style={font=\small\bfseries,color=freedodark},grid=major,grid style={gray!18}]
\nextgroupplot[title={(a) Synthetic restarted response fit},ylabel={Response amplitude},
 legend style={font=\scriptsize,draw=none,fill=white!90,at={(.03,.05)},anchor=south west}]
\addplot[black,line width=1.1pt] coordinates {(0.000,0.000000) (0.020,-0.048342) (0.040,-0.096687) (0.060,-0.144934) (0.080,-0.192985) (0.100,-0.240742) (0.120,-0.288109) (0.140,-0.334993) (0.160,-0.381303) (0.180,-0.426949) (0.200,-0.471844) (0.220,-0.515904) (0.240,-0.559046) (0.260,-0.601192) (0.280,-0.642265) (0.300,-0.682192) (0.320,-0.720901) (0.340,-0.758324) (0.360,-0.794399) (0.380,-0.829062) (0.400,-0.862256) (0.420,-0.893926) (0.440,-0.924021) (0.460,-0.952492) (0.480,-0.979294) (0.500,-1.004386) (0.520,-1.027731) (0.540,-1.049295) (0.560,-1.069045) (0.580,-1.086956) (0.600,-1.103002) (0.620,-1.117165) (0.640,-1.129427) (0.660,-1.139775) (0.680,-1.148198) (0.700,-1.154692) (0.720,-1.159251) (0.740,-1.161877) (0.760,-1.162574) (0.780,-1.161347) (0.800,-1.158208) (0.820,-1.153168) (0.840,-1.146245) (0.860,-1.137458) (0.880,-1.126828) (0.900,-1.114382) (0.920,-1.100145) (0.940,-1.084150) (0.960,-1.066429) (0.980,-1.047018) (1.000,-1.025954) (1.020,-1.003279) (1.040,-0.979035) (1.060,-0.953266) (1.080,-0.926019) (1.100,-0.897344) (1.120,-0.867290) (1.140,-0.835909) (1.160,-0.803256) (1.180,-0.769386) (1.200,-0.734356) (1.220,-0.698223) (1.240,-0.661047) (1.260,-0.622888) (1.280,-0.583807) (1.300,-0.543866) (1.320,-0.503127) (1.340,-0.461654) (1.360,-0.419511) (1.380,-0.376761) (1.400,-0.333469) (1.420,-0.289699) (1.440,-0.245516) (1.460,-0.200984) (1.480,-0.156168) (1.500,-0.111131) (1.520,-0.065937) (1.540,-0.020650) (1.560,0.024668) (1.580,0.069956) (1.600,0.115152) (1.620,0.160195) (1.640,0.205028) (1.660,0.249590) (1.680,0.293825) (1.700,0.337678) (1.720,0.381092) (1.740,0.424016) (1.760,0.466396) (1.780,0.508182) (1.800,0.549326) (1.820,0.589779) (1.840,0.629497) (1.860,0.668435) (1.880,0.706550) (1.900,0.743803) (1.920,0.780155) (1.940,0.815568) (1.960,0.850008) (1.980,0.883443) (2.000,0.915840)};
\addlegendentry{Target tail $2\operatorname{Re}[C_p(r+s)-C_p(r)]$}
\addplot[eventorange,dashed,line width=1.1pt] coordinates {(0.000,0.000000) (0.020,-0.032061) (0.040,-0.066449) (0.060,-0.102903) (0.080,-0.141170) (0.100,-0.181002) (0.120,-0.222160) (0.140,-0.264413) (0.160,-0.307534) (0.180,-0.351307) (0.200,-0.395522) (0.220,-0.439976) (0.240,-0.484475) (0.260,-0.528832) (0.280,-0.572868) (0.300,-0.616412) (0.320,-0.659299) (0.340,-0.701375) (0.360,-0.742492) (0.380,-0.782510) (0.400,-0.821296) (0.420,-0.858726) (0.440,-0.894684) (0.460,-0.929062) (0.480,-0.961757) (0.500,-0.992677) (0.520,-1.021735) (0.540,-1.048854) (0.560,-1.073963) (0.580,-1.096997) (0.600,-1.117900) (0.620,-1.136624) (0.640,-1.153126) (0.660,-1.167371) (0.680,-1.179330) (0.700,-1.188981) (0.720,-1.196309) (0.740,-1.201305) (0.760,-1.203966) (0.780,-1.204294) (0.800,-1.202300) (0.820,-1.197997) (0.840,-1.191407) (0.860,-1.182554) (0.880,-1.171469) (0.900,-1.158188) (0.920,-1.142751) (0.940,-1.125205) (0.960,-1.105597) (0.980,-1.083982) (1.000,-1.060418) (1.020,-1.034966) (1.040,-1.007690) (1.060,-0.978660) (1.080,-0.947947) (1.100,-0.915625) (1.120,-0.881772) (1.140,-0.846468) (1.160,-0.809794) (1.180,-0.771835) (1.200,-0.732678) (1.220,-0.692409) (1.240,-0.651118) (1.260,-0.608896) (1.280,-0.565835) (1.300,-0.522026) (1.320,-0.477564) (1.340,-0.432543) (1.360,-0.387055) (1.380,-0.341195) (1.400,-0.295058) (1.420,-0.248737) (1.440,-0.202325) (1.460,-0.155916) (1.480,-0.109600) (1.500,-0.063470) (1.520,-0.017614) (1.540,0.027878) (1.560,0.072919) (1.580,0.117424) (1.600,0.161309) (1.620,0.204492) (1.640,0.246892) (1.660,0.288432) (1.680,0.329035) (1.700,0.368629) (1.720,0.407143) (1.740,0.444507) (1.760,0.480657) (1.780,0.515528) (1.800,0.549062) (1.820,0.581199) (1.840,0.611886) (1.860,0.641070) (1.880,0.668705) (1.900,0.694743) (1.920,0.719143) (1.940,0.741867) (1.960,0.762877) (1.980,0.782142) (2.000,0.799632)};
\addlegendentry{Original basis ($5.62\times 10^{-2}$)}
\addplot[freedoblue,dotted,line width=1.3pt] coordinates {(0.000,0.000000) (0.020,-0.048342) (0.040,-0.096687) (0.060,-0.144934) (0.080,-0.192985) (0.100,-0.240742) (0.120,-0.288109) (0.140,-0.334993) (0.160,-0.381303) (0.180,-0.426949) (0.200,-0.471844) (0.220,-0.515904) (0.240,-0.559046) (0.260,-0.601192) (0.280,-0.642265) (0.300,-0.682192) (0.320,-0.720901) (0.340,-0.758324) (0.360,-0.794399) (0.380,-0.829062) (0.400,-0.862256) (0.420,-0.893926) (0.440,-0.924021) (0.460,-0.952492) (0.480,-0.979294) (0.500,-1.004386) (0.520,-1.027731) (0.540,-1.049295) (0.560,-1.069045) (0.580,-1.086956) (0.600,-1.103002) (0.620,-1.117165) (0.640,-1.129427) (0.660,-1.139775) (0.680,-1.148198) (0.700,-1.154692) (0.720,-1.159251) (0.740,-1.161877) (0.760,-1.162574) (0.780,-1.161347) (0.800,-1.158208) (0.820,-1.153168) (0.840,-1.146245) (0.860,-1.137458) (0.880,-1.126828) (0.900,-1.114382) (0.920,-1.100145) (0.940,-1.084150) (0.960,-1.066429) (0.980,-1.047018) (1.000,-1.025954) (1.020,-1.003279) (1.040,-0.979035) (1.060,-0.953266) (1.080,-0.926019) (1.100,-0.897344) (1.120,-0.867290) (1.140,-0.835909) (1.160,-0.803256) (1.180,-0.769386) (1.200,-0.734356) (1.220,-0.698223) (1.240,-0.661047) (1.260,-0.622888) (1.280,-0.583807) (1.300,-0.543866) (1.320,-0.503127) (1.340,-0.461654) (1.360,-0.419511) (1.380,-0.376761) (1.400,-0.333469) (1.420,-0.289699) (1.440,-0.245516) (1.460,-0.200984) (1.480,-0.156168) (1.500,-0.111131) (1.520,-0.065937) (1.540,-0.020650) (1.560,0.024668) (1.580,0.069956) (1.600,0.115152) (1.620,0.160195) (1.640,0.205028) (1.660,0.249590) (1.680,0.293825) (1.700,0.337678) (1.720,0.381092) (1.740,0.424016) (1.760,0.466396) (1.780,0.508182) (1.800,0.549326) (1.820,0.589779) (1.840,0.629497) (1.860,0.668435) (1.880,0.706550) (1.900,0.743803) (1.920,0.780155) (1.940,0.815568) (1.960,0.850008) (1.980,0.883443) (2.000,0.915840)};
\addlegendentry{Extended with $\Psi_\kappa(s)$ ($3.80\times 10^{-16}$)}
\nextgroupplot[title={(b) Original-basis fit residual},ylabel={Fit minus target}]
\addplot[eventorange,line width=1.0pt] coordinates {(0.000,0.000000) (0.020,0.016281) (0.040,0.030238) (0.060,0.042031) (0.080,0.051816) (0.100,0.059740) (0.120,0.065949) (0.140,0.070581) (0.160,0.073769) (0.180,0.075642) (0.200,0.076322) (0.220,0.075927) (0.240,0.074571) (0.260,0.072360) (0.280,0.069397) (0.300,0.065780) (0.320,0.061601) (0.340,0.056949) (0.360,0.051907) (0.380,0.046552) (0.400,0.040960) (0.420,0.035200) (0.440,0.029336) (0.460,0.023430) (0.480,0.017537) (0.500,0.011710) (0.520,0.005996) (0.540,0.000440) (0.560,-0.004917) (0.580,-0.010041) (0.600,-0.014898) (0.620,-0.019459) (0.640,-0.023699) (0.660,-0.027596) (0.680,-0.031132) (0.700,-0.034290) (0.720,-0.037058) (0.740,-0.039428) (0.760,-0.041392) (0.780,-0.042947) (0.800,-0.044092) (0.820,-0.044829) (0.840,-0.045161) (0.860,-0.045096) (0.880,-0.044640) (0.900,-0.043806) (0.920,-0.042606) (0.940,-0.041055) (0.960,-0.039168) (0.980,-0.036964) (1.000,-0.034464) (1.020,-0.031686) (1.040,-0.028655) (1.060,-0.025394) (1.080,-0.021928) (1.100,-0.018282) (1.120,-0.014483) (1.140,-0.010559) (1.160,-0.006538) (1.180,-0.002449) (1.200,0.001679) (1.220,0.005815) (1.240,0.009929) (1.260,0.013992) (1.280,0.017972) (1.300,0.021839) (1.320,0.025563) (1.340,0.029112) (1.360,0.032456) (1.380,0.035565) (1.400,0.038411) (1.420,0.040962) (1.440,0.043191) (1.460,0.045068) (1.480,0.046568) (1.500,0.047661) (1.520,0.048323) (1.540,0.048528) (1.560,0.048251) (1.580,0.047468) (1.600,0.046157) (1.620,0.044296) (1.640,0.041864) (1.660,0.038842) (1.680,0.035210) (1.700,0.030952) (1.720,0.026051) (1.740,0.020492) (1.760,0.014261) (1.780,0.007346) (1.800,-0.000264) (1.820,-0.008581) (1.840,-0.017612) (1.860,-0.027365) (1.880,-0.037846) (1.900,-0.049060) (1.920,-0.061011) (1.940,-0.073701) (1.960,-0.087131) (1.980,-0.101301) (2.000,-0.116208)};
\end{groupplot}
\end{tikzpicture}
\caption{Deterministic basis diagnostic verifying Eq.~\eqref{eq:closure} ($\kappa=1, p=-0.35+2\mathrm{i}, r=0.6$). Fitting on even grid points and evaluating on odd grid points over $s \in [0, 2]$ yields a relative $\ell_2$ error of $5.6210\times 10^{-2}$ for the original three-column basis and $3.80\times 10^{-16}$ after adjoining the missing direction $\Psi_\kappa(s)$.}
\label{fig:basis_diagnostic}
\end{figure}

Figure~\ref{fig:basis_diagnostic} validates Eq.~\eqref{eq:closure} numerically: fitting $2\operatorname{Re}[C_p(r+s)-C_p(r)]$ for $\kappa=1, p=-0.35+2\mathrm{i}, r=0.6$ via FP64 least squares leaves a held-out relative $\ell_2$ residual of $5.6210\times 10^{-2}$ on the original basis, which drops to machine precision ($3.80\times 10^{-16}$) upon adjoining $\Psi_\kappa(s)$. Furthermore, even when $\Psi_\kappa(s)$ closes the anchored exponential function space, re-initializing at $\widehat{\mathbf{q}}(t_r)$ resets the action-prefix GRU hidden state $\mathbf{h}_k$. This analysis clarifies that the flow loss $\mathcal{L}_{\mathsf{flow}}$ in Eq.~\eqref{eq:flow} acts as an approximate regularizer rather than an exact semigroup closure identity.

\subsection{Koopman resolvent connection and finite-pool decision regret bound}\label{sec:koopman_regret}

\noindent\textbf{Conditional connection to Koopman generator resolvents.}
For an autonomous dynamical system with flow map $S^t$, the Koopman semigroup $\{U^t\}_{t\ge 0}$ acts on observable functions by $(U^t f)(x) = f(S^t x)$. Under strong continuity and exponential growth bounds, the infinitesimal generator $\mathcal{G}$ admits the Laplace resolvent representation \citep{susuki2026resolvents}:
\begin{equation}\label{eq:koopman_resolvent}
R(s;\, \mathcal{G}) = (sI - \mathcal{G})^{-1} = \int_0^\infty e^{-st} U^t\,\mathrm{d}t,
\end{equation}
in a convergent right half-plane. If a vector of observables $\boldsymbol{\psi}(x)$ spans a finite-dimensional $\mathcal{G}$-invariant subspace such that $\mathcal{G}\boldsymbol{\psi} = \mathbf{A}\boldsymbol{\psi}$ with a linear readout $f = \mathbf{C}\boldsymbol{\psi}$, the transformed trajectory equals $\mathbf{C}(sI - \mathbf{A})^{-1}\boldsymbol{\psi}(x_0)$, yielding a rational pole-residue expansion. In \ours{}, while Eq.~\eqref{eq:full_laplace} exhibits a rational pole-residue structure for each event response, the coefficients $(\mathbf{v}_k, \mathbf{b}_k)$ depend nonlinearly on $\mathbf{q}_0$ and the causal prefix $\mathbf{u}_{0:k}$, and the visual readout $D_\theta$ is a nonlinear residual MLP that does not commute with the Laplace integral $\mathcal{L}\{\cdot\}$. Consequently, the learned poles $\{p_m\}_{m=1}^M$ serve as expressive continuous-time basis modes for latent action responses rather than identified eigenvalues of a global linear Koopman operator.

\noindent\textbf{Finite-pool decision regret bound.}
To connect embedding prediction accuracy to CEM planning quality over a finite candidate action pool $\mathcal{U}$, let $\mathbf{e}_j(\mathbf{u})$ denote the true future visual embedding of candidate $\mathbf{u} \in \mathcal{U}$, $\widehat{\mathbf{e}}_j(\mathbf{u}) = D_\theta(\widehat{\mathbf{q}}(t_j))$ denote the model prediction, and $\mathbf{e}_{g,j}$ denote the target goal embedding with nonnegative horizon weights $w_j$ summing to $1$ (for terminal-only cost, $w_K = 1$). Suppose that across all candidates $\mathbf{u} \in \mathcal{U}$, the true goal distance satisfies $\|\mathbf{e}_j(\mathbf{u}) - \mathbf{e}_{g,j}\|_2 \le R_j$ and the model prediction error satisfies $\|\widehat{\mathbf{e}}_j(\mathbf{u}) - \mathbf{e}_j(\mathbf{u})\|_2 \le \epsilon_j$. Expanding the squared Euclidean cost $J_e(\mathbf{u}) = \sum_j w_j \|\mathbf{e}_j(\mathbf{u}) - \mathbf{e}_{g,j}\|_2^2$ and its predicted counterpart $\widehat{J}(\mathbf{u})$ yields the uniform cost perturbation bound:
\begin{equation}\label{eq:regret_bound}
\bigl|\widehat{J}(\mathbf{u}) - J_e(\mathbf{u})\bigr| \le \epsilon_J := \sum_j w_j\bigl(2R_j\epsilon_j + \epsilon_j^2\bigr)
\quad\Longrightarrow\quad
J_e(\widehat{\mathbf{u}}) \le J_e(\mathbf{u}^\star) + 2\epsilon_J + \eta,
\end{equation}
where $\widehat{\mathbf{u}} \in \mathcal{U}$ is an $\eta$-optimal plan selected by the planner under $\widehat{J}$ and $\mathbf{u}^\star \in \arg\min_{\mathbf{u}\in\mathcal{U}} J_e(\mathbf{u})$ is the true optimal candidate in the pool. Equation~\eqref{eq:regret_bound} shows that reducing terminal embedding prediction error $\epsilon_K$ directly tightens the suboptimality upper bound of the selected action plan in representation space.

\section{System engineering and standalone \texttt{pole\_worldmodel} package}\label{sec:engineering}

To ensure industrial-grade reproducibility and deployment readiness at Beijing Freedo Technology Co., Ltd., we package the balanced V3 model (retaining the exported class name \texttt{CRLeWMV3} for checkpoint lineage compatibility) into a standalone, zero-workspace-dependency Python software stack \texttt{pole\_worldmodel}.

\begin{table}[t!]
\centering
\caption{Complete architectural specifications and training/planning hyperparameters of \ours{} (Balanced-V3).}
\label{tab:model_spec}
\fontsize{8}{10}\selectfont
\setlength{\tabcolsep}{3.0pt}
\begin{tabular}{ll|ll}
\toprule
Architecture Module & \ours{} Specification & Training \& Planning Parameter & \ours{} Specification \\
\midrule
Visual encoder $E_\theta$ & ViT-Tiny/14, $224\times224$, from scratch & Training / Split seeds & Seeds $17, 29$ / Split seed $3072$ \\
Feature projector & MLP $192\to2048\to192$, BatchNorm & Optimizer \& Peak LR & AdamW, LR $5\times10^{-5}$, WD $10^{-3}$ \\
History \& Readout $D$ & $D=192$, GRU history + residual MLP & Batch / Epochs / Clip & Batch $128$ ($2\times 64$ DDP) / $20$ / $1.0$ \\
Prefix GRU \& Gate width & $D_{\mathsf{dyn}} = 512$ hidden units & Cold-history augmentation & Prob.\ $0.25$ (repeat $\mathbf{e}_0$, zero $\mathbf{U}$) \\
Conjugate pole pairs $M$ & $M=25$ pairs ($50$ poles), $\kappa=1.0$ & Sliced QQ regularization & $P=1024$ projections, all $J=8$ frames \\
Decay \& Freq.\ floors & $d_{\min}=0.05$, $\omega_{\min}=0.1$ & Precision boundaries & BF16 vision; FP32 heads; FP64 basis \\
History / Target / Macro & $L=3$ frames / $5$ macro / $b=5$ raw & CEM search budget & $300$ candidates, $30$ iters, $30$ elites \\
Macro / Micro clock & $\Delta t = 0.25$ / $\delta = 0.05$ ($K=25$ steps) & Total trainable parameters & $18,134,578$ ($244$ state tensors) \\
\bottomrule
\end{tabular}
\end{table}

\noindent\textbf{Three-tier numerical precision and initialization contract.}
As summarized in Table~\ref{tab:model_spec}, \ours{} comprises $18,134,578$ trainable parameters across $244$ state tensors and enforces strict precision isolation across three computational tiers: (i) the ViT-Tiny/14 visual backbone and 2048-wide BatchNorm projector support BF16 automatic mixed precision (AMP); (ii) the history GRU, action-prefix GRU, free-response head, and innovation-gate networks explicitly exit AMP autocast to compute drift and residue coefficients in full FP32, with terminal linear layers zero-initialized so initial predictions start stationary at $\mathbf{q}_0$; and (iii) initial pole decay excesses are log-spaced in $[0.25, 2.5]$ above $d_{\min}=0.05$ and frequency excesses are linearly spaced in $[0.4, 8.0]$ above $\omega_{\min}=0.1$, with all basis evaluations and event summations performed in full FP64 / \texttt{complex128} precision.

\begin{table}[t!]
\centering
\caption{Comparison of learning-rate schedules, optimizer step budgets, and deterministic BatchNorm calibration protocols in our standalone \texttt{pole\_worldmodel} software stack.}
\label{tab:schedule_bn}
\fontsize{8}{10}\selectfont
\setlength{\tabcolsep}{2.8pt}
\begin{tabular}{lcccc}
\toprule
Training / Calibration Protocol & Epoch Range & Optimizer Updates & Warmup & Protocol Specification \& Verification \\
\midrule
Historical Segment 1 (\texttt{balanced\_9}) & Epochs 1--9 & $102,510$ & $1,025$ & Cosine decay to 0; $11,390$ steps/epoch (Cube) \\
Historical Segment 2 (LR Restart) & Epochs 10--20 & $+125,290$ ($227,800$) & $1,252$ & Preserves Adam moments; resets cosine clock \\
Direct 20-Epoch Cosine (\texttt{direct20}) & Epochs 1--20 & $227,800$ & $2,278$ & Uninterrupted single cosine cycle from scratch \\
\midrule
Train-Only BN Recalibration & Post-hoc frozen & $128 \times 64 = 8,192$ win. & N/A & Seed $314159$ train permutation; cumulative mean \\
Offline Fixed-Window Validation & Post-hoc frozen & $64 \times 64 = 4,096$ win. & N/A & Seed $99173$ held-out windows; FP64 SSE sum \\
\bottomrule
\end{tabular}
\end{table}

\noindent\textbf{Learning-rate restart schedule and deterministic BN calibration.}
Table~\ref{tab:schedule_bn} documents the two distinct 20-epoch training protocols supported by \texttt{pole\_worldmodel}: the uninterrupted 20-epoch cosine schedule used in our multi-environment two-seed benchmark (\S\ref{sec:experiments}) and the historical 9-epoch plus 11-epoch learning-rate restart protocol used for the released Cube V3 artifact. In addition, \texttt{pole\_worldmodel.training.calibrate} implements a strictly train-only BatchNorm recalibration utility that resets projector running buffers (\texttt{running\_mean}, \texttt{running\_var}, \texttt{num\_batches\_tracked}) and accumulates exact cumulative averages (\texttt{momentum=None}) over $8,192$ training windows sampled without replacement under seed $314159$, reproducing all $244$ model-state tensors bitwise.

\section{Experiments}\label{sec:experiments}

In this section, we evaluate \ours{} through the following four research questions:
\begin{itemize}[leftmargin=*]
  \item \textbf{RQ1 (Multi-Environment Control Performance):} Does \ours{} improve image-goal planning success across diverse continuous control and navigation tasks compared with autoregressive and direct world-model baselines?
  \item \textbf{RQ2 (Response-Structure Attribution):} To what extent does the early control advantage arise from the action-response factorization itself versus the specific choice of neural or analytic pole time basis?
  \item \textbf{RQ3 (Complete-Solve Planning Efficiency):} Does analytic pole-basis evaluation maintain low end-to-end CEM planning latency when visual encoding and all 30 iterative search steps are included?
  \item \textbf{RQ4 (Training Dynamics, Checkpoint Selection, and Standalone Replay):} How stably does \ours{} evolve across all 20 training epochs and random seeds, and does the standalone package verify bitwise reproducibility?
\end{itemize}

\subsection{Experiment setup}\label{sec:exp_setup}

\noindent\textbf{Evaluation benchmarks and dataset splits.}
We evaluate on four offline image-goal control environments: \textbf{Cube} (3D robotic block manipulation from OGBench \citep{park2025ogbench}), \textbf{TwoRoom} (2D visual doorway navigation), \textbf{PushT} (planar non-prehensile T-block pushing), and \textbf{Reacher} (two-link continuous robotic arm reaching). Table~\ref{tab:dataset_counts} reports the total trajectories, training windows, validation windows, and 20-epoch optimizer update counts for each environment under the shared 90/10 random window split (seed $3072$, global batch size $128$). All compared methods share full-dataset action normalization statistics and identical 300-task evaluation pools per environment ($6$ evaluation seeds $\times 50$ start/goal pairs, where each goal image is taken $25$ raw steps ahead of the initial state).

\begin{table}[t!]
\centering
\caption{Dataset statistics, 90/10 window partitions (split seed 3072), and 20-epoch gradient update budgets per training seed.}
\label{tab:dataset_counts}
\fontsize{8}{10}\selectfont
\setlength{\tabcolsep}{13.5pt}
\begin{tabular}{lrrrr}
\toprule
Environment & Total Episodes & Training Windows (90\%) & Validation Windows (10\%) & 20-Epoch Optimizer Updates \\
\midrule
Cube & 10,000 & 1,458,000 & 162,000 & 227,800 \\
Reacher & 10,000 & 1,458,000 & 162,000 & 227,800 \\
PushT & 18,685 & 1,447,219 & 160,802 & 226,120 \\
TwoRoom & 10,000 & 477,786 & 53,087 & 74,640 \\
\bottomrule
\end{tabular}
\end{table}

\noindent\textbf{Baselines and closed-loop planning protocol.}
We compare \ours{} against four representative joint-embedding world models evaluated under matched planning budgets: (1) \textbf{LeWM} \citep{maes2026lewm}, an autoregressive one-step latent predictor; (2) \textbf{Fast-LeWM} \citep{gao2026fast}, a direct action-prefix predictor; (3) \textbf{Fast-LeWM + consistency} \citep{gao2026fast}, which augments direct prediction with composition consistency; and (4) \textbf{QQWorld} \citep{yu2026qq}, which integrates sliced quantile-quantile embedding regularization. Every planner runs CEM with $300$ candidates, $30$ iterations, $30$ elites, and a planning horizon of $5$ macro actions ($K = 25$ raw controls). Each episode executes up to $25$ raw controls before replanning within a total budget of $50$ environment steps: the first CEM solve uses the initial RGB frame (cold-start padding), and the second solve at step $25$ uses the real RGB observations at steps $15, 20, 25$ and the intervening executed actions. Success is scored by the native geometric predicates within $50$ steps: Cube 3D Euclidean position error $\le 0.04\,\mathrm{m}$; Reacher per-joint angle error $< 0.05\,\mathrm{rad}$; PushT joint agent/block $xy$ error $< 20$ with wrapped block orientation error $< \pi/9$; and TwoRoom goal distance $< 16\,\mathrm{px}$.

\subsection{(RQ1) Multi-environment goal-conditioned control}\label{sec:rq1}

\begin{table}[t!]
\centering
\caption{Goal-reaching control success rates (\%) across four visual planning benchmarks (300 tasks per environment and checkpoint). \ours{} reports mean $\pm$ sample standard deviation across training seeds 17 and 29 at fixed epoch 20 (primary) and under validation-loss checkpoint selection.}
\label{tab:main}
\fontsize{8}{10}\selectfont
\setlength{\tabcolsep}{9.8pt}
\begin{tabular}{lrrrrr}
\toprule
Method & Cube & TwoRoom & PushT & Reacher & Macro Mean \\
\midrule
LeWM \citep{maes2026lewm} & 69.67 & 83.67 & 84.00 & 78.00 & 78.83 \\
Fast-LeWM \citep{gao2026fast} & 73.33 & 90.33 & 87.33 & 82.00 & 83.25 \\
Fast-LeWM + consistency \citep{gao2026fast} & 76.67 & 90.33 & 86.33 & 80.67 & 83.50 \\
QQWorld \citep{yu2026qq} & 71.00 & 92.33 & \textbf{91.67} & 84.00 & 84.75 \\
\midrule
\rowcolor{tabhighlight}
\ours{} (Fixed Epoch 20, ours) & $\mathbf{88.67}_{\pm0.47}$ & $\mathbf{97.67}_{\pm0.94}$ & $82.67_{\pm1.41}$ & $\mathbf{94.33}_{\pm0.00}$ & $\mathbf{90.83}_{\pm0.47}$ \\
\rowcolor{tabhighlight}
\ours{} (Validation-Selected, ours) & $\mathbf{89.50}_{\pm0.71}$ & $\mathbf{97.67}_{\pm0.94}$ & $82.67_{\pm1.41}$ & $\mathbf{95.00}_{\pm0.94}$ & $\mathbf{91.21}_{\pm1.00}$ \\
\bottomrule
\end{tabular}
\end{table}

Table~\ref{tab:main} reports the four-environment control comparison. We have the following observations:
\begin{itemize}[leftmargin=*]
  \item \ours{} \textbf{achieves the highest macro-average success rate and leads on three of the four environments.} At the fixed epoch-20 endpoint, \ours{} reaches $90.83\% \pm 0.47\%$ macro-average success, outperforming QQWorld ($84.75\%$) by $+6.08$ percentage points, Fast-LeWM + consistency ($83.50\%$) by $+7.33$ points, Fast-LeWM ($83.25\%$) by $+7.58$ points, and autoregressive LeWM ($78.83\%$) by $+12.00$ points. Across individual benchmarks, \ours{} improves over the strongest baseline on Cube ($88.67\%$ vs.\ $76.67\%$, $+12.00$ points), TwoRoom ($97.67\%$ vs.\ $92.33\%$, $+5.34$ points), and Reacher ($94.33\%$ vs.\ $84.00\%$, $+10.33$ points). We attribute these broad gains to the action-response factorization, which shares temporal dynamics across horizons while preserving exact initial-state anchoring by construction.
  \item \textbf{Validation-loss checkpoint selection further improves mean control success to $91.21\% \pm 1.00\%$.} Selecting checkpoints solely by minimum validation embedding prediction loss $\mathcal{L}_{\mathsf{pred}}$ within each training run (without inspecting control outcomes) shifts Seed 17 on Cube and Reacher from epoch 20 to epoch 19 while keeping all other six checkpoints at epoch 20, lifting Cube to $89.50\% \pm 0.71\%$ and Reacher to $95.00\% \pm 0.94\%$.
  \item \textbf{PushT exhibits a task-specific trade-off against QQWorld.} On PushT, \ours{} achieves $82.67\% \pm 1.41\%$ compared with $84.00\%$ for LeWM, $87.33\%$ for Fast-LeWM, and $91.67\%$ for QQWorld. Planar non-prehensile pushing involves discontinuous stick-slip contact transitions, whereas Cube, TwoRoom, and Reacher are dominated by continuous articulated motion or spatial navigation where smooth compensated exponential responses align closely with the underlying trajectories.
\end{itemize}

\subsection{(RQ2) Response-structure attribution in early training}\label{sec:ablation}

To disentangle the structural effect of action-response factorization Eq.~\eqref{eq:factorization} from the specific choice of analytic pole basis Eq.~\eqref{eq:basis} and from restart flow regularization Eq.~\eqref{eq:flow}, we conduct a controlled internal head comparison during the first training epoch on Cube ($11,390$ gradient updates, evaluated every $0.1$ epoch / $1,139$ updates with $\lambda_{\mathsf{flow}} = 0$, training seed $17$, global batch $128$, and shared train-only BN recalibration over $8,192$ windows):
\begin{enumerate}
  \item \textbf{Head R (Generic Direct Predictor):} maps the conditioned state and action-prefix representation directly to target future states without decomposing trajectories into free and delayed event responses.
  \item \textbf{Head A (Neural-Time Response Head):} retains the exact free-plus-delayed action-response factorization of Eq.~\eqref{eq:factorization} and the bias-free innovation gate of Eq.~\eqref{eq:gate}, but replaces the $2M$ analytic pole transient channels with a learned bounded neural time basis $\mathbf{B}_\theta(t) = \rho(t)^2\bigl[1-\rho(t)\bigr]\mathbf{g}_\theta(\rho(t)) \in \R^{2M}$ over compact warped time $\rho(t) \in [0, 1)$, which enforces zero onset value and slope via the factor $\rho(t)^2$ and vanishing long-time transients via $1-\rho(t)$.
  \item \textbf{Head C (\ours{} Analytic Pole Response Head):} uses the real anchor $A(t)$ and 25 conjugate pairs of second-order compensated exponentials $C_{p_m}(t)$ in Eq.~\eqref{eq:basis}--\eqref{eq:response}.
\end{enumerate}

\begin{figure}[t!]
\centering
{\small\textcolor{gray}{\rule[.5ex]{.55cm}{1.0pt}} R: Generic Direct Head\quad
\textcolor{poleblue}{\rule[.5ex]{.55cm}{1.2pt}} A: Neural-Time Response Head\quad
\textcolor{eventorange}{\rule[.5ex]{.55cm}{1.2pt}} C: \ours{} Pole Response Head\par}
\vspace{3pt}
\begin{minipage}[c]{.48\linewidth}
\centering
\includegraphics[width=\linewidth]{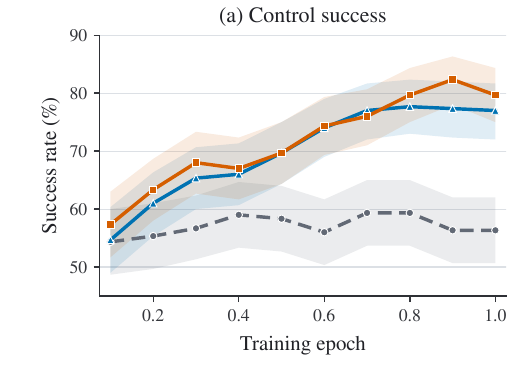}
\end{minipage}\hfill
\begin{minipage}[c]{.48\linewidth}
\centering
\begin{tikzpicture}
\begin{axis}[width=.78\linewidth,height=4.3cm,xbar,
 xmin=0,xmax=28,xtick={0,10,20},
 ytick={0,1,2},yticklabels={C $-$ A (Pole vs.\ Neural),C $-$ R (Pole vs.\ Direct),A $-$ R (Neural vs.\ Direct)},
 ymin=-.6,ymax=2.6,bar width=11pt,
 axis x line*=bottom,axis y line*=left,
 xlabel={Epoch-1 success diff.\ (pp)},
 title={(b) Epoch-1.0 head contrasts},
 title style={font=\small\bfseries,color=freedodark},tick label style={font=\scriptsize},label style={font=\footnotesize},
 xmajorgrids=true,grid style={gray!15},
 nodes near coords,point meta=x,
 every node near coord/.append style={font=\scriptsize\bfseries,anchor=west,/pgf/number format/fixed,/pgf/number format/precision=2}]
\addplot[fill=eventorange!75,draw=none] coordinates {(2.67,0)};
\addplot[fill=poleblue!75,draw=none] coordinates {(23.34,1) (20.67,2)};
\end{axis}
\end{tikzpicture}
\end{minipage}
\caption{First-epoch Cube response-head comparison every $0.1$ epoch ($1,139$ updates) under $\lambda_{\mathsf{flow}}=0$ and training seed 17. Left: control success rate (\%) across 300 tasks, where shaded bands denote pointwise 95\% task-bootstrap intervals over 10,000 resamples. Right: percentage-point contrasts at epoch 1.0.}
\label{fig:early}
\end{figure}
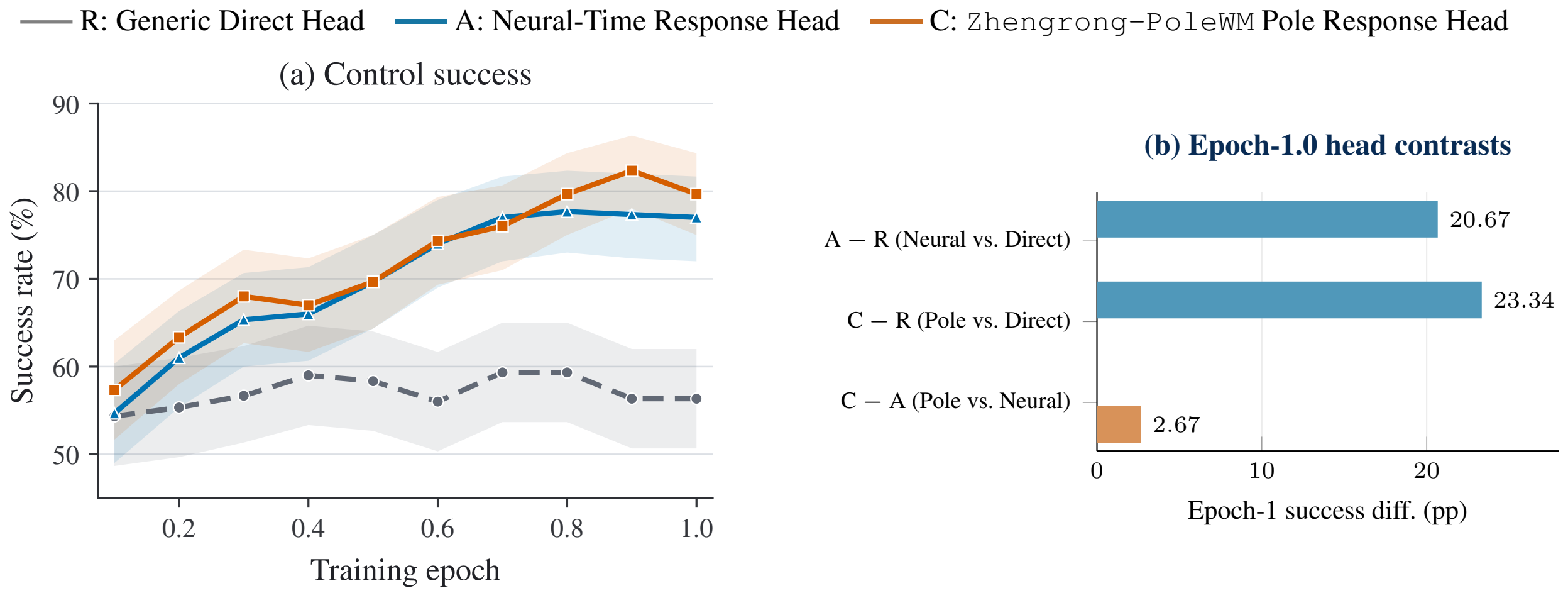

\begin{table}[t!]
\centering
\caption{Cube control success rates (\%) and pairwise percentage-point contrasts at two fixed early checkpoints (epochs 0.5 and 1.0) with $\lambda_{\mathsf{flow}}=0$, training seed 17, and shared 8,192-window train-only BN recalibration.}
\label{tab:heads}
\fontsize{8}{10}\selectfont
\setlength{\tabcolsep}{3.5pt}
\begin{tabular}{lrrrccc}
\toprule
Training Progress & R: Generic Direct & A: Neural Resp. & C: \ours{} Pole & A $-$ R Gain & C $-$ R Gain & C $-$ A Contrast \\
\midrule
Epoch 0.5 ($5,695$ updates) & 58.33\% & 69.67\% & \textbf{69.67\%} & $+11.34$ pp & $+11.34$ pp & $+0.00$ pp \\
\rowcolor{tabhighlight}
Epoch 1.0 ($11,390$ updates) & 56.33\% & 77.00\% & \textbf{79.67\%} & $\mathbf{+20.67}$ \textbf{pp} & $\mathbf{+23.34}$ \textbf{pp} & $\mathbf{+2.67}$ \textbf{pp} \\
\bottomrule
\end{tabular}
\end{table}

Figure~\ref{fig:early} and Table~\ref{tab:heads} summarize the early-training head comparison. We have the following observations:
\begin{itemize}[leftmargin=*]
  \item \textbf{Action-response organization accounts for the primary early-learning improvement.} At epoch 0.5 ($5,695$ updates), both response heads reach $69.67\%$ success compared with $58.33\%$ for the generic direct head R ($+11.34$ percentage points). By epoch 1.0 ($11,390$ updates), the neural response head A reaches $77.00\%$ ($+20.67$ points over R) and the pole response head C reaches $79.67\%$ ($+23.34$ points over R). Because all three variants use $\lambda_{\mathsf{flow}} = 0$, this $>20$-point gain occurs independently of restart flow regularization and demonstrates that decomposing predictions into free and delayed action responses provides a strong sample-efficiency inductive bias.
  \item \textbf{The analytic pole basis achieves the highest epoch-1 accuracy while providing closed-form structural guarantees.} Comparing the two response heads at epoch 1.0, the analytic pole head C outperforms the neural response head A by $+2.67$ percentage points ($79.67\%$ vs.\ $77.00\%$) while requiring only $50$ learnable pole scalars instead of an auxiliary time-basis MLP and guaranteeing the exact Laplace and asymptotic properties of \S\ref{sec:structure}.
\end{itemize}

\subsection{(RQ3) Complete-solve planning efficiency}\label{sec:timing}

\begin{table}[t!]
\centering
\caption{Single-frame complete CEM planning latency medians (ms) on an exclusive NVIDIA A100-SXM4-80GB GPU (4 warmups + 9 timed solves per environment; 300 candidates $\times$ 30 iterations, first visual encoding included).}
\label{tab:timing}
\fontsize{8}{10}\selectfont
\setlength{\tabcolsep}{4.8pt}
\begin{tabular}{lrrrrrc}
\toprule
Method & Cube & TwoRoom & PushT & Reacher & Task Average & Relative to Fast-LeWM \\
\midrule
LeWM (Autoregressive Rollout) & 548\,ms & 551\,ms & 549\,ms & 550\,ms & 549.5\,ms & $+497.3\%$ ($6.0\times$ slower) \\
Fast-LeWM (Direct Action-Prefix) & 91\,ms & 92\,ms & 91\,ms & 94\,ms & 92.0\,ms & Baseline ($1.00\times$) \\
\rowcolor{tabhighlight}
\ours{} (Analytic Pole, ours) & \textbf{87\,ms} & \textbf{74\,ms} & \textbf{75\,ms} & \textbf{74\,ms} & \textbf{77.5\,ms} & $\mathbf{-15.8\%}$ ($\mathbf{7.1\times}$ \textbf{vs.\ LeWM}) \\
\bottomrule
\end{tabular}
\end{table}

We benchmark end-to-end CEM planning latency under a standardized single-frame protocol on an exclusive NVIDIA A100-SXM4-80GB GPU (Table~\ref{tab:timing}). Each timed call receives one current image and one goal image with a fresh visual cache, brackets the entire 30-iteration, 300-candidate CEM search with \texttt{torch.cuda.synchronize()}, and includes host-device transfers, initial ViT encoding, trajectory evaluation, and action return. We have the following observations:
\begin{itemize}[leftmargin=*]
  \item \ours{} \textbf{completes a full 30-iteration CEM search in $74\text{--}87\,\mathrm{ms}$, outperforming both autoregressive and direct baselines.} Across the four tasks, \ours{} achieves a mean task median latency of $77.5\,\mathrm{ms}$, delivering a $7.1\times$ speedup over autoregressive LeWM ($549.5\,\mathrm{ms}$) and reducing mean latency by $15.8\%$ relative to Fast-LeWM ($92.0\,\mathrm{ms}$), with per-task latency reductions of $4.4\%$ on Cube, $19.6\%$ on TwoRoom, $17.6\%$ on PushT, and $21.3\%$ on Reacher. Even with FP64 / \texttt{complex128} analytic basis construction and summation, caching the free response and evaluating closed-form pole bases keeps full-search latency well within real-time control budgets.
\end{itemize}

\subsection{(RQ4) Training dynamics, checkpoint selection, and standalone replay}\label{sec:rq4}

\begin{figure}[t!]
\centering
\begin{tikzpicture}
\begin{groupplot}[group style={group size=2 by 2,horizontal sep=1.15cm,vertical sep=1.15cm},
 width=.455\linewidth,height=3.85cm,
 xmin=1,xmax=20,ymin=0,ymax=100,xtick={1,5,10,15,20},ytick={0,50,100},
 xlabel={Training epoch},ylabel={Control success (\%)},
 tick label style={font=\scriptsize},label style={font=\small},title style={font=\small\bfseries,color=freedodark},
 grid=major,grid style={gray!15},legend style={font=\scriptsize,draw=none,fill=white!90,at={(.03,.05)},anchor=south west}]
\nextgroupplot[title=Cube (3D Manipulation),xlabel={}]
\addplot[poleblue,line width=1.1pt] coordinates {(1,76.00) (2,78.33) (3,80.33) (4,81.00) (5,83.33) (6,85.33) (7,88.67) (8,87.67) (9,87.67) (10,89.00) (11,90.33) (12,88.67) (13,88.67) (14,89.33) (15,89.00) (16,88.67) (17,89.67) (18,89.33) (19,89.33) (20,88.33)};
\addlegendentry{Training Seed 17}
\addplot[eventorange,line width=1.1pt,dashed] coordinates {(1,65.33) (2,82.33) (3,80.00) (4,81.33) (5,80.67) (6,84.33) (7,86.00) (8,85.33) (9,89.33) (10,88.00) (11,87.67) (12,90.33) (13,89.67) (14,88.33) (15,89.00) (16,87.33) (17,87.33) (18,89.67) (19,89.33) (20,89.00)};
\addlegendentry{Training Seed 29}
\nextgroupplot[title=Reacher (Continuous Articulated Control),xlabel={},ylabel={}]
\addplot[poleblue,line width=1.1pt] coordinates {(1,35.33) (2,33.33) (3,36.33) (4,65.33) (5,30.00) (6,79.33) (7,78.67) (8,83.00) (9,88.33) (10,80.67) (11,92.67) (12,89.00) (13,93.67) (14,92.00) (15,93.00) (16,92.33) (17,94.00) (18,94.00) (19,96.33) (20,94.33)};
\addplot[eventorange,line width=1.1pt,dashed] coordinates {(1,19.00) (2,22.00) (3,39.67) (4,67.33) (5,78.00) (6,63.00) (7,86.67) (8,81.67) (9,84.00) (10,81.33) (11,92.33) (12,90.00) (13,94.67) (14,94.00) (15,95.00) (16,93.33) (17,94.67) (18,94.67) (19,93.00) (20,94.33)};
\nextgroupplot[title=PushT (Planar Non-Prehensile Pushing)]
\addplot[poleblue,line width=1.1pt] coordinates {(1,36.00) (2,63.67) (3,68.67) (4,75.33) (5,75.00) (6,76.33) (7,75.67) (8,81.67) (9,82.00) (10,83.33) (11,81.33) (12,82.00) (13,83.00) (14,83.33) (15,84.67) (16,83.00) (17,85.33) (18,82.33) (19,84.33) (20,83.67)};
\addplot[eventorange,line width=1.1pt,dashed] coordinates {(1,44.67) (2,62.33) (3,68.33) (4,69.33) (5,75.67) (6,75.67) (7,76.00) (8,80.00) (9,82.00) (10,81.67) (11,80.00) (12,81.33) (13,79.67) (14,81.67) (15,80.33) (16,83.00) (17,80.33) (18,81.00) (19,80.00) (20,81.67)};
\nextgroupplot[title=TwoRoom (Visual Doorway Navigation),ylabel={}]
\addplot[poleblue,line width=1.1pt] coordinates {(1,46.33) (2,67.00) (3,74.00) (4,91.67) (5,88.00) (6,98.33) (7,99.67) (8,99.33) (9,96.33) (10,99.33) (11,96.33) (12,98.33) (13,99.33) (14,98.33) (15,98.67) (16,98.67) (17,98.33) (18,97.67) (19,98.33) (20,98.33)};
\addplot[eventorange,line width=1.1pt,dashed] coordinates {(1,57.67) (2,55.67) (3,90.67) (4,79.33) (5,85.67) (6,88.00) (7,98.00) (8,99.00) (9,97.67) (10,97.67) (11,98.00) (12,98.00) (13,97.67) (14,98.00) (15,96.33) (16,97.00) (17,97.00) (18,98.00) (19,97.33) (20,97.00)};
\end{groupplot}
\end{tikzpicture}
\caption{Epoch-by-epoch control success trajectories (\%) of \ours{} across all 20 training epochs for training seeds 17 and 29 ($4 \times 2 \times 20 = 160$ checkpoints, 300 tasks per point, $48,000$ evaluated episodes in total).}
\label{fig:fullcurves}
\end{figure}
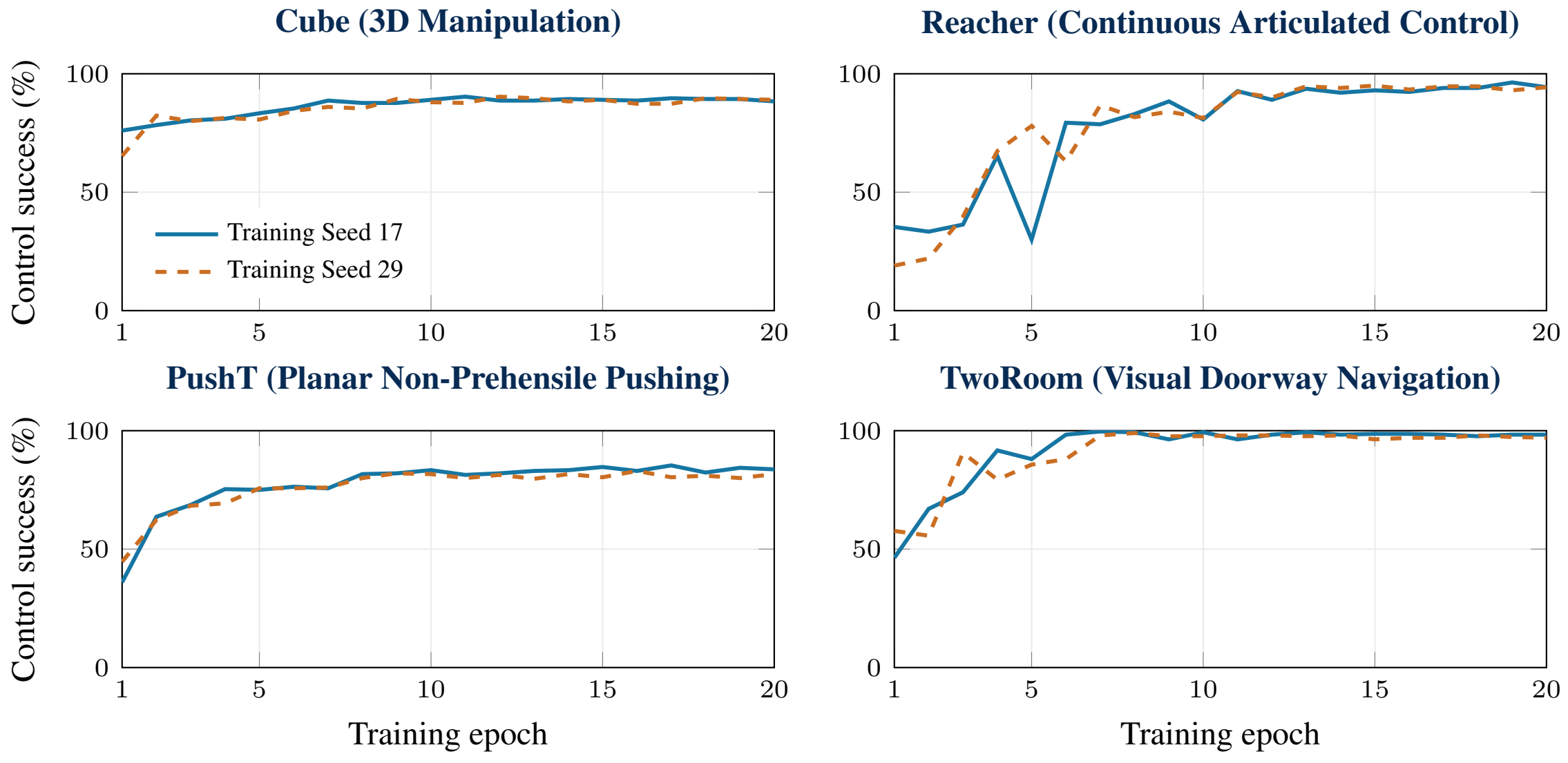

\begin{table}[t!]
\centering
\caption{Standalone \texttt{pole\_worldmodel} package verification on the modern SWM 0.1.1 Cube protocol (single-frame + goal RGB input, 300 tasks across 6 evaluation seeds, NVIDIA EGL rendering backend).}
\label{tab:standalone_replay}
\fontsize{8}{10}\selectfont
\setlength{\tabcolsep}{5.2pt}
\begin{tabular}{llccc}
\toprule
Evaluated Checkpoint Artifact & BN Statistics Treatment & CEM Iters & Solved (/300) & Success (\%) \\
\midrule
Official LeWM Checkpoint \citep{maes2026lewm} & Original checkpoint BN buffers & 30 & 209 / 300 & 69.67\% \\
\ours{} Balanced-V3 Epoch 10 & Train-only 8,192-win.\ BN calibration & 10 & 236 / 300 & 78.67\% \\
\ours{} Balanced-V3 Epoch 10 & Train-only 8,192-win.\ BN calibration & 30 & 248 / 300 & 82.67\% \\
\ours{} Balanced-V3 Epoch 20 & Train-only 8,192-win.\ BN calibration & 10 & 249 / 300 & 83.00\% \\
\rowcolor{tabhighlight}
\ours{} Balanced-V3 Epoch 20 (Replay) & Train-only 8,192-win.\ BN calibration & 30 & \textbf{269 / 300} & \textbf{89.67\%} \\
\bottomrule
\end{tabular}
\end{table}

Figure~\ref{fig:fullcurves} plots the complete 20-epoch control progression across both training seeds ($160$ checkpoints, $48,000$ formal episodes; full numerical tables in \S\ref{app:results}), and Table~\ref{tab:standalone_replay} reports the standalone package verification across CEM iteration budgets. We have the following observations:
\begin{itemize}[leftmargin=*]
  \item \textbf{Both training seeds converge smoothly and consistently across all four environments.} On TwoRoom, Reacher, and Cube, both seeds exceed $85\%$ success by epoch 10 and plateau tightly between epochs 15 and 20 with inter-seed standard deviations below $1.0\%$ at epoch 20 ($0.47\%$ on Cube, $0.94\%$ on TwoRoom, and $0.00\%$ on Reacher where both seeds achieve exactly $283/300$).
  \item \textbf{The extracted \texttt{pole\_worldmodel} package achieves bitwise state reproduction and 100\% task-outcome agreement.} As shown in Table~\ref{tab:standalone_replay}, even with only $10$ CEM iterations at epoch 10 ($78.67\%$), \ours{} surpasses the 30-iteration official LeWM checkpoint ($69.67\%$) by $+9.00$ percentage points. At epoch 20 with 30 CEM iterations, the standalone package completes $269/300$ tasks ($89.67\%$), matching the reference evaluation on every single one of the 300 tasks while reproducing all 244 model-state tensors bitwise after train-only BN recalibration.
\end{itemize}

\section{Limitations and conclusion}\label{sec:conclusion}

\noindent\textbf{Scope and methodological boundaries.}
First, on contact-rich planar pushing (PushT), QQWorld achieves higher success ($91.67\%$ vs.\ $82.67\%$), indicating that smooth exponential pole modes can be further enriched with hybrid contact-event primitives for non-prehensile manipulation. Second, the four-environment comparison uses two independent training seeds under a shared 90/10 overlapping-window split and common 300-task development pools, which evaluates matched in-domain control rather than trajectory-held-out zero-shot generalization. Third, the $74\text{--}87\,\mathrm{ms}$ CEM timing benchmark uses a standardized single-frame input protocol, whereas the $90.83\%$ control evaluation uses three-frame online observation history; the two experiments separately characterize planning latency and closed-loop control quality. Finally, Proposition~\ref{prop:bounded} establishes open-loop boundedness for a fixed finite plan rather than closed-loop Lyapunov stability under repeated nonlinear receding-horizon control replanning.

\noindent\textbf{Conclusion.}
We presented \ours{}, an action-response visual world model developed at Beijing Freedo Technology Co., Ltd. By factoring direct latent trajectory predictions into an initial-state anchor, a free response, and causally delayed action-innovation responses evaluated over a shared bank of second-order compensated conjugate poles, \ours{} decouples nonlinear prefix conditioning from continuous-time evolution. Our theoretical analysis proves exact initial anchoring, action non-anticipation, and finite-plan boundedness, while identifying the exact exponential-polynomial direction $\Psi_\kappa(s)$ that separates single-response boundedness from restart closure. Empirically, \ours{} achieves $90.83\%$ macro-average control success across four visual planning benchmarks ($+6.08$ percentage points over the strongest baseline), demonstrates a $+23.34$-point early-learning advantage over a generic direct predictor, and completes full 30-iteration CEM planning in $74\text{--}87\,\mathrm{ms}$ on an A100 GPU.

\newpage
\appendix

\section{Mathematical proofs and theoretical derivations}\label{app:proofs}

\subsection{Proof of Proposition~\ref{prop:causal} (anchoring, continuity, and non-anticipation)}
From Eq.~\eqref{eq:basis}, direct evaluation at $t = 0$ gives $A(0) = \frac{1-1}{\kappa} = 0$ and $C_p(0) = 1 - 1 - 0 = 0$. Differentiating with respect to $t$ yields $A'(t) = e^{-\kappa t}$ and $C'_p(t) = p e^{pt} + \kappa e^{-\kappa t} - (p+\kappa)(1-\kappa t)e^{-\kappa t}$, so $A'(0) = 1$ and $C'_p(0) = p + \kappa - (p+\kappa) = 0$. At query time $t = 0$, the free response $\boldsymbol{\phi}(0; \mathbf{v}_{\mathsf{f}}, \mathbf{b}_{\mathsf{f}}) = \mathbf{0}$ and the initial event response $\boldsymbol{\phi}(0; \mathbf{v}_0, \mathbf{b}_0) = \mathbf{0}$, while all subsequent events $k \ge 1$ have $\tau_k = k\delta > 0$ and are inactive ($\mathbf{1}_{0 \ge \tau_k} = 0$). Substituting into Eq.~\eqref{eq:response} proves $\widehat{\mathbf{q}}(0) = \mathbf{q}_0$. Because each event term $\mathbf{1}_{t \ge \tau_k}\boldsymbol{\phi}(t-\tau_k; \mathbf{v}_k, \mathbf{b}_k)$ vanishes at $t = \tau_k$ and is infinitely differentiable on $(\tau_k, \infty)$, their finite sum $\widehat{\mathbf{q}}(t)$ is continuous on $[0, \infty)$, with the right derivative jumping by $\mathbf{v}_k$ at $t = \tau_k$.

For any query time $t \ge 0$, any event with $\tau_k > t$ is zeroed by the indicator $\mathbf{1}_{t \ge \tau_k}$, and any event with $\tau_k = t$ contributes $\boldsymbol{\phi}(0; \mathbf{v}_k, \mathbf{b}_k) = \mathbf{0}$. Every active event ($\tau_k < t$) computes its hidden state $\mathbf{h}_k$ via a unidirectional GRU over $\mathbf{u}_{0:k}$ and its innovation gate from $\mathbf{u}_k - \mathbf{u}_{k-1}$. Consequently, modifying any action at or after time $t$ leaves $\widehat{\mathbf{q}}(t)$ strictly unchanged. Finally, when normalized-zero actions are appended after $\mathbf{u}_{K-1}$, the first transition to zero at $\tau_K = K\delta$ shares the exact same prefix $\mathbf{u}_{0:K}$. Every subsequent appended step $k > K$ has $\mathbf{u}_k = \mathbf{u}_{k-1} = \mathbf{0}$, so $\mathbf{u}_k - \mathbf{u}_{k-1} = \mathbf{0}$. Because $\mathbf{W}_\theta$ in Eq.~\eqref{eq:gate} has no bias, $\mathbf{W}_\theta(\mathbf{u}_k - \mathbf{u}_{k-1}) = \mathbf{0} \implies \boldsymbol{\eta}_k = \mathbf{0} \implies (\mathbf{v}_k, \mathbf{b}_k) = (\mathbf{0}, \mathbf{0})$, leaving $\widehat{\mathbf{q}}(t)$ identical for all $t \ge 0$. $\square$

\subsection{Proof of Proposition~\ref{prop:bounded} and Laplace transform identities}
For all $t \ge 0$ and $\kappa > 0$, we have $0 \le A(t) = \frac{1-e^{-\kappa t}}{\kappa} \le \kappa^{-1}$. For any pole $p$ in the open left half-plane ($\operatorname{Re}p < 0$), applying the triangle inequality to $C_p(t)$ in Eq.~\eqref{eq:basis} gives:
\begin{equation}\label{eq:cp_bound}
|C_p(t)| \le \bigl|e^{pt}\bigr| + e^{-\kappa t} + |p + \kappa|\,t e^{-\kappa t}
= e^{(\operatorname{Re}p)t} + e^{-\kappa t} + |p + \kappa|\,t e^{-\kappa t}
\le 2 + \frac{|p + \kappa|}{e\kappa},
\end{equation}
where we used $e^{(\operatorname{Re}p)t} \le 1$, $e^{-\kappa t} \le 1$, and $\max_{t \ge 0} t e^{-\kappa t} = (e\kappa)^{-1}$ attained at $t = \kappa^{-1}$. Applying the triangle inequality to Eq.~\eqref{eq:response} across the free response and the $K+1$ events directly yields the uniform upper bound in Eq.~\eqref{eq:explicit_bound}. As $t \to \infty$, since $\operatorname{Re}p_m \le -d_{\min} < 0$ and $\kappa > 0$, we have $\lim_{t\to\infty} e^{p_m(t-\tau_k)} = 0$, $\lim_{t\to\infty} e^{-\kappa(t-\tau_k)} = 0$, and $\lim_{t\to\infty} (t-\tau_k)e^{-\kappa(t-\tau_k)} = 0$, so $\lim_{t\to\infty} C_{p_m}(t-\tau_k) = 0$ and $\lim_{t\to\infty} A(t-\tau_k) = \kappa^{-1}$ for every $k \in \{0, \ldots, K\}$. Substituting these limits into Eq.~\eqref{eq:response} establishes Eq.~\eqref{eq:limit}. $\square$

To verify the Laplace transform identities in Eq.~\eqref{eq:laplace_basis} for $\operatorname{Re}s > 0$, direct integration yields $\mathcal{L}\{A\}(s) = \kappa^{-1}\bigl(s^{-1} - (s+\kappa)^{-1}\bigr) = \frac{1}{s(s+\kappa)}$ and:
\begin{equation}
G_p(s) = \mathcal{L}\{C_p\}(s) = \frac{1}{s-p} - \frac{1}{s+\kappa} - \frac{p+\kappa}{(s+\kappa)^2}
= \frac{(s+\kappa)^2 - (s-p)(s+\kappa) - (p+\kappa)(s-p)}{(s-p)(s+\kappa)^2}
= \frac{(p+\kappa)^2}{(s-p)(s+\kappa)^2}.
\end{equation}
Applying the Laplace time-shift theorem $\mathcal{L}\{\mathbf{1}_{t\ge\tau_k}f(t-\tau_k)\}(s) = e^{-s\tau_k}\mathcal{L}\{f\}(s)$ to each event response yields Eq.~\eqref{eq:full_laplace}. $\square$

\subsection{Proof of the re-anchoring identity Eq.~\eqref{eq:closure} and span restriction}
Let $r > 0$ and $s \ge 0$. Expanding the left-hand side of Eq.~\eqref{eq:closure} using Eq.~\eqref{eq:basis} gives:
\begin{equation}\label{eq:lhs_expand}
C_p(r+s) - C_p(r) = e^{pr}e^{ps} - e^{-\kappa r}e^{-\kappa s} - (p+\kappa)(r+s)e^{-\kappa r}e^{-\kappa s} - C_p(r).
\end{equation}
On the right-hand side of Eq.~\eqref{eq:closure}, substituting $C_p(s) = e^{ps} - e^{-\kappa s} - (p+\kappa)se^{-\kappa s}$, $A(s) = \frac{1-e^{-\kappa s}}{\kappa}$, $\Psi_\kappa(s) = \frac{1-e^{-\kappa s}}{\kappa} - se^{-\kappa s}$, and $C'_p(r) = pe^{pr} + \kappa e^{-\kappa r} - (p+\kappa)(1-\kappa r)e^{-\kappa r}$, we collect coefficients with respect to the four basis functions $\{e^{ps},\, se^{-\kappa s},\, 1,\, e^{-\kappa s}\}$:
\begin{itemize}[leftmargin=*]
  \item Coefficient of $e^{ps}$: $e^{pr}$, matching Eq.~\eqref{eq:lhs_expand}.
  \item Coefficient of $se^{-\kappa s}$: $-e^{pr}(p+\kappa) - (p+\kappa)(e^{-\kappa r} - e^{pr}) = -(p+\kappa)e^{-\kappa r}$, matching Eq.~\eqref{eq:lhs_expand}.
  \item Constant term $1$: $\kappa^{-1}\bigl[C'_p(r) + (p+\kappa)(e^{-\kappa r} - e^{pr})\bigr] = \kappa^{-1}\bigl[\kappa e^{-\kappa r} + \kappa(p+\kappa)re^{-\kappa r} - \kappa e^{pr}\bigr] = -C_p(r)$, matching the expansion in Eq.~\eqref{eq:lhs_expand}.
  \item Coefficient of $e^{-\kappa s}$: $-e^{pr} - \kappa^{-1}\bigl[C'_p(r) + (p+\kappa)(e^{-\kappa r} - e^{pr})\bigr] = -e^{pr} + C_p(r) = -e^{-\kappa r} - (p+\kappa)re^{-\kappa r}$, matching the expansion in Eq.~\eqref{eq:lhs_expand}.
\end{itemize}
Because all four coefficients match identically, Eq.~\eqref{eq:closure} holds for all $r, s \ge 0$. Since the exponential-polynomial functions $\{1, e^{-\kappa s}, \allowbreak se^{-\kappa s}, \allowbreak e^{p_1 s}, e^{\overline{p_1}s}, \ldots, e^{p_M s}, e^{\overline{p_M}s}\}$ are linearly independent over $\C$ when $\{p_m\}$ are distinct and nonreal ($\omega_m \ge \omega_{\min} > 0$), any linear combination of $\{A(s), \operatorname{Re}C_{p_m}(s), \allowbreak\operatorname{Im}C_{p_m}(s)\}$ that eliminates all oscillatory terms $e^{p_m s}$ must set all $C_{p_m}(s)$ weights to zero, leaving only a scalar multiple of $A(s) = \frac{1-e^{-\kappa s}}{\kappa}$, which cannot generate the $se^{-\kappa s}$ term in $\Psi_\kappa(s)$. Adjoining $\Psi_\kappa(s)$ completes the shift-invariant anchored exponential-polynomial subspace $\bigl\{f \in \operatorname{span}_{\R}\{1, e^{-\kappa s}, \allowbreak se^{-\kappa s}, \allowbreak\operatorname{Re}e^{p_m s}, \allowbreak\operatorname{Im}e^{p_m s}\} :\allowbreak f(0) = 0\bigr\}$. $\square$

\subsection{State-scale invariance analysis}
In Eq.~\eqref{eq:loss}, $\mathcal{L}_{\mathsf{QQ}}$ regularizes the visual embeddings $\mathbf{e}_j$ directly rather than the latent states $\mathbf{q}_j$. At the functional level, rescaling the state coordinates by a nonzero scalar $a \neq 0$ via $\mathbf{q} \mapsto a\mathbf{q}$, $F_\theta(t; \mathbf{q}, \mathbf{u}) \mapsto a F_\theta(t; \mathbf{q}/a, \mathbf{u})$, and $D_\theta(\mathbf{q}) \mapsto D_\theta(\mathbf{q}/a)$ leaves the decoded embedding predictions $\widehat{\mathbf{e}}(t_j) = D_\theta(\widehat{\mathbf{q}}(t_j))$ and $\mathcal{L}_{\mathsf{pred}}, \mathcal{L}_{\mathsf{read}}, \mathcal{L}_{\mathsf{QQ}}$ invariant while scaling $\mathcal{L}_{\mathsf{state}}$ and $\mathcal{L}_{\mathsf{flow}}$ by $a^2$. In practice, the residual connection $\mathbf{q}_0 = \mathbf{e}_0 + \Delta\mathbf{q}_0$ in $H_\theta$ and $D_\theta(\mathbf{q}) = \mathbf{q} + \Delta D_\theta(\mathbf{q})$ together with AdamW weight decay anchor the scale of $\mathbf{q}$ to the unit-variance QQ-regularized visual embedding space $\mathbf{e}$.

\section{Complete per-seed and epoch-by-epoch control results}\label{app:results}

Table~\ref{tab:seed_raw} lists the raw task success counts (out of $300$) for training seeds 17 and 29 under both fixed epoch 20 and validation-loss checkpoint selection. Table~\ref{tab:full_20epochs} provides the complete numerical control success rates (\%) across all 20 training epochs for every environment and training seed ($160$ checkpoints, $48,000$ evaluated control episodes), corresponding to the curves in Figure~\ref{fig:fullcurves}.

\begin{table}[h!]
\centering
\caption{Raw task success counts (out of 300 tasks per checkpoint) for training seeds 17 and 29 under fixed epoch 20 and within-run validation prediction-loss selection.}
\label{tab:seed_raw}
\fontsize{8}{10}\selectfont
\setlength{\tabcolsep}{12.5pt}
\begin{tabular}{llccc}
\toprule
Environment & Training Seed & Fixed Epoch 20 Successes & Validation-Selected Epoch & Selected-Epoch Successes \\
\midrule
Cube & 17 & 265/300 & 19 & 268/300 \\
Cube & 29 & 267/300 & 20 & 267/300 \\
Reacher & 17 & 283/300 & 19 & 289/300 \\
Reacher & 29 & 283/300 & 20 & 283/300 \\
PushT & 17 & 251/300 & 20 & 251/300 \\
PushT & 29 & 245/300 & 20 & 245/300 \\
TwoRoom & 17 & 295/300 & 20 & 295/300 \\
TwoRoom & 29 & 291/300 & 20 & 291/300 \\
\bottomrule
\end{tabular}
\end{table}

\begin{table}[h!]
\centering
\caption{Complete epoch-by-epoch control success rates (\%) of \ours{} across all 20 training epochs for training seeds 17 and 29 across Cube, Reacher, PushT, and TwoRoom ($300$ evaluation tasks per cell; $48,000$ total episodes).}
\label{tab:full_20epochs}
\fontsize{8}{10}\selectfont
\setlength{\tabcolsep}{6.0pt}
\begin{tabular}{c|cc|cc|cc|cc|c}
\toprule
\multirow{2}{*}{Epoch} & \multicolumn{2}{c|}{Cube (\%)} & \multicolumn{2}{c|}{Reacher (\%)} & \multicolumn{2}{c|}{PushT (\%)} & \multicolumn{2}{c|}{TwoRoom (\%)} & \multirow{2}{*}{Two-Seed Macro Mean (\%)} \\
 & Seed 17 & Seed 29 & Seed 17 & Seed 29 & Seed 17 & Seed 29 & Seed 17 & Seed 29 & \\
\midrule
1  & 76.00 & 65.33 & 40.33 & 23.00 & 36.00 & 53.33 & 50.67 & 69.00 & 51.71 \\
2  & 78.33 & 82.33 & 36.33 & 36.00 & 66.33 & 57.67 & 60.67 & 94.33 & 64.00 \\
3  & 80.33 & 80.00 & 48.00 & 35.67 & 66.67 & 66.33 & 75.67 & 86.33 & 67.38 \\
4  & 81.00 & 81.33 & 72.00 & 57.33 & 75.67 & 72.33 & 90.00 & 94.67 & 78.04 \\
5  & 83.33 & 80.67 & 40.67 & 66.67 & 76.00 & 75.00 & 97.00 & 96.33 & 76.96 \\
6  & 85.33 & 84.33 & 83.33 & 80.67 & 75.67 & 75.00 & 94.67 & 97.67 & 84.58 \\
7  & 88.67 & 86.00 & 90.00 & 88.67 & 77.67 & 78.33 & 95.33 & 97.33 & 87.75 \\
8  & 87.67 & 85.33 & 92.67 & 83.67 & 82.67 & 83.00 & 97.00 & 97.00 & 88.63 \\
9  & 87.67 & 89.33 & 90.67 & 92.33 & 82.33 & 83.00 & 95.67 & 98.33 & 89.92 \\
10 & 89.00 & 88.00 & 89.67 & 91.33 & 83.33 & 83.33 & 96.67 & 96.33 & 89.71 \\
11 & 90.33 & 87.67 & 94.67 & 94.67 & 84.67 & 82.67 & 98.00 & 97.33 & 91.25 \\
12 & 88.67 & 90.33 & 91.67 & 94.00 & 82.67 & 83.67 & 98.00 & 97.33 & 90.79 \\
13 & 88.67 & 89.67 & 93.33 & 95.67 & 81.33 & 84.67 & 96.00 & 97.00 & 90.79 \\
14 & 89.33 & 88.33 & 91.00 & 95.33 & 81.33 & 84.00 & 97.67 & 97.67 & 90.58 \\
15 & 89.00 & 89.00 & 92.33 & 94.67 & 83.00 & 81.33 & 97.00 & 97.67 & 90.50 \\
16 & 88.67 & 87.33 & 92.00 & 95.00 & 84.33 & 83.00 & 97.00 & 97.67 & 90.63 \\
17 & 89.67 & 87.33 & 94.00 & 94.33 & 83.67 & 81.33 & 98.67 & 98.00 & 90.88 \\
18 & 89.33 & 89.67 & 96.33 & 95.33 & 83.67 & 80.67 & 98.33 & 97.33 & 91.33 \\
19 & 89.33 & 89.33 & 96.33 & 94.33 & 84.33 & 81.67 & 98.00 & 96.67 & 91.25 \\
\rowcolor{tabhighlight}
20 & \textbf{88.33} & \textbf{89.00} & \textbf{94.33} & \textbf{94.33} & \textbf{83.67} & \textbf{81.67} & \textbf{98.33} & \textbf{97.00} & \textbf{90.83} \\
\bottomrule
\end{tabular}
\end{table}

\bibliographystyle{iclr2027_conference}
\bibliography{references}

\end{document}